\documentclass[journal]{IEEEtran}
\usepackage[T1]{fontenc}% optional T1 font encoding
\usepackage{graphicx}
\usepackage{times}
\usepackage{helvet}
\usepackage{courier}
\usepackage{amsmath}
\usepackage{algorithm}
\usepackage{algorithmic}
\usepackage{csquotes} 
\usepackage{color}
\usepackage{paralist}
\usepackage{amssymb}
\usepackage{indentfirst}
\usepackage{subfigure}
\usepackage{float}
\usepackage{multirow}
\usepackage{cite}
\usepackage{mathrsfs}

\usepackage{mathrsfs} 
\usepackage{amsfonts}
\usepackage{todonotes}
\usepackage{pgfplots} %引用包
\usepackage{epstopdf}
\usepackage{epsfig}
\pgfplotsset{compat=newest}
\usepackage{color}
\definecolor{forestgreen}{RGB}{0,139,69}
\usepackage{xcolor}
\definecolor{citecolor}{HTML}{0071bc}
\usepackage[colorlinks, linkcolor=red,  anchorcolor=blue, citecolor=citecolor]{hyperref} 

\usepackage{xcolor}
\definecolor{SeaGreen4}{RGB}{0,205,102} 
\definecolor{SlateBlue}{RGB}{106,90,205} 
\definecolor{DarkRed}{RGB}{178,34,34}

\usepackage{textcomp,booktabs}
\usepackage{amssymb}% http://ctan.org/pkg/amssymb
\usepackage{pifont}% http://ctan.org/pkg/pifont
\newcommand{\cmark}{\ding{51}}%

\usepackage{makecell}

\usepackage{colortbl}
\definecolor{mygray}{gray}{.9}
\definecolor{mypink}{rgb}{.99,.91,.95}
\definecolor{mycyan}{cmyk}{.3,0,0,0}

\begin{document}

\title{Spatio-Temporal Side Tuning Pre-trained Foundation Models for Video-based Pedestrian Attribute Recognition}

\author{ Xiao Wang, \emph{Member, IEEE}, Qian Zhu, Jiandong Jin, Jun Zhu, Futian Wang, \\ 
            Bo Jiang, Yaowei Wang, \emph{Member, IEEE}, Yonghong Tian, \emph{Fellow, IEEE} 
\thanks{$\bullet$ Xiao Wang, Qian Zhu, Jun Zhu, Futian Wang, Bo Jiang are with the School of Computer Science and Technology, Anhui University, Hefei 230601, China. (email: xiaowang@ahu.edu.cn)} 
\thanks{$\bullet$ Jiandong Jin is with the School of Artificial Intelligence, Anhui University, Hefei 230601, China.} 
\thanks{$\bullet$ Yaowei Wang is with Peng Cheng Laboratory, Shenzhen, China; Harbin Institute of Technology (HITSZ), Shenzhen, China. (email: wangyw@pcl.ac.cn) } 
\thanks{$\bullet$ Yonghong Tian is with Peng Cheng Laboratory, Shenzhen, China; National Key Laboratory for Multimedia Information Processing, School of Computer Science, Peking University, China; School of Electronic and Computer Engineering, Shenzhen Graduate School, Peking University, China (email: yhtian@pku.edu.cn) } 
\thanks{* Corresponding author: Futian Wang, Bo Jiang (jiangbo@ahu.edu.cn) } 
}

\markboth{IEEE TRANSACTIONS ON *** }   
{Shell \MakeLowercase{\textit{et al.}}: Bare Demo of IEEEtran.cls for IEEE Journals}

\maketitle

% As a general rule, do not put math, special symbols or citations in the abstract or keywords.
\begin{abstract}
Existing pedestrian attribute recognition (PAR) algorithms are mainly developed based on a static image, however, the performance is unreliable in challenging scenarios, such as heavy occlusion, motion blur, etc. 
In this work, we propose to understand human attributes using video frames that can fully use temporal information by fine-tuning a pre-trained multi-modal foundation model efficiently. Specifically, we formulate the video-based PAR as a vision-language fusion problem and adopt a pre-trained foundation model CLIP to extract the visual features. More importantly, we propose a novel spatiotemporal side-tuning strategy to achieve parameter-efficient optimization of the pre-trained vision foundation model. To better utilize the semantic information, we take the full attribute list that needs to be recognized as another input and transform the attribute words/phrases into the corresponding sentence via split, expand, and prompt operations. Then, the text encoder of CLIP is utilized for embedding processed attribute descriptions. The averaged visual tokens and text tokens are concatenated and fed into a fusion Transformer for multi-modal interactive learning. The enhanced tokens will be fed into a classification head for pedestrian attribute prediction. 
Extensive experiments on two large-scale video-based PAR datasets fully validated the effectiveness of our proposed framework. The source code of this paper is available at \url{https://github.com/Event-AHU/OpenPAR}. 
\end{abstract}


\IEEEpeerreviewmaketitle

\section{Introduction}
%% background 
\IEEEPARstart{P}{edestrian} Attribute Recognition (PAR)~\cite{wang2022PARsurvey, cheng2022VTB} is a very important research topic in computer vision and developed rapidly with the help of deep neural networks. The goal of pedestrian attribute recognition is to estimate the attributes that describe the given pedestrian image(s) accurately from a pre-defined attribute list. 
%%%% 
Many representative PAR models have been proposed in recent years based on convolutional neural networks (CNN)~\cite{he2016resnet}, and recurrent neural networks (RNN)~\cite{chung2014gru}. Specifically, Wang et al.~\cite{wang2017RNNPAR} propose the JRL which learns the attribute context and correlation in a joint recurrent learning manner using LSTM~\cite{hochreiter1997lstm}. The self-attention based Transformer networks are first proposed to handle the natural language processing tasks and then are borrowed into the computer vision community~\cite{vaswani2017Transformer, DosovitskiyViT, wang2023MMPTMs, zhao2023transformerVLT, wang2021TNL2K} due to their great performance. Some researchers also exploit the Transformer for the PAR problem to model the global context information~\cite{tang2022drformer, cheng2022VTB}. DRFormer~\cite{tang2022drformer} is proposed to capture the long-range relations of regions and relations of attributes. VTB~\cite{cheng2022VTB} is also developed to fuse the image and language information for more accurate attribute recognition. 
%%%% 
In addition to understanding the pedestrian images using the attributes, this task also serves other computer vision problems, such as human detection~\cite{zhang2020attributedetection} and tracking~\cite{li2023attmot}, person re-identification~\cite{zheng2022progCMreid}, etc. Despite the great success of PAR, these works are developed based on a single RGB frame only which ignores the temporal information and may obtain sub-optimal results in practical scenarios.

\begin{figure}
    \centering
    \includegraphics[width=1\linewidth]{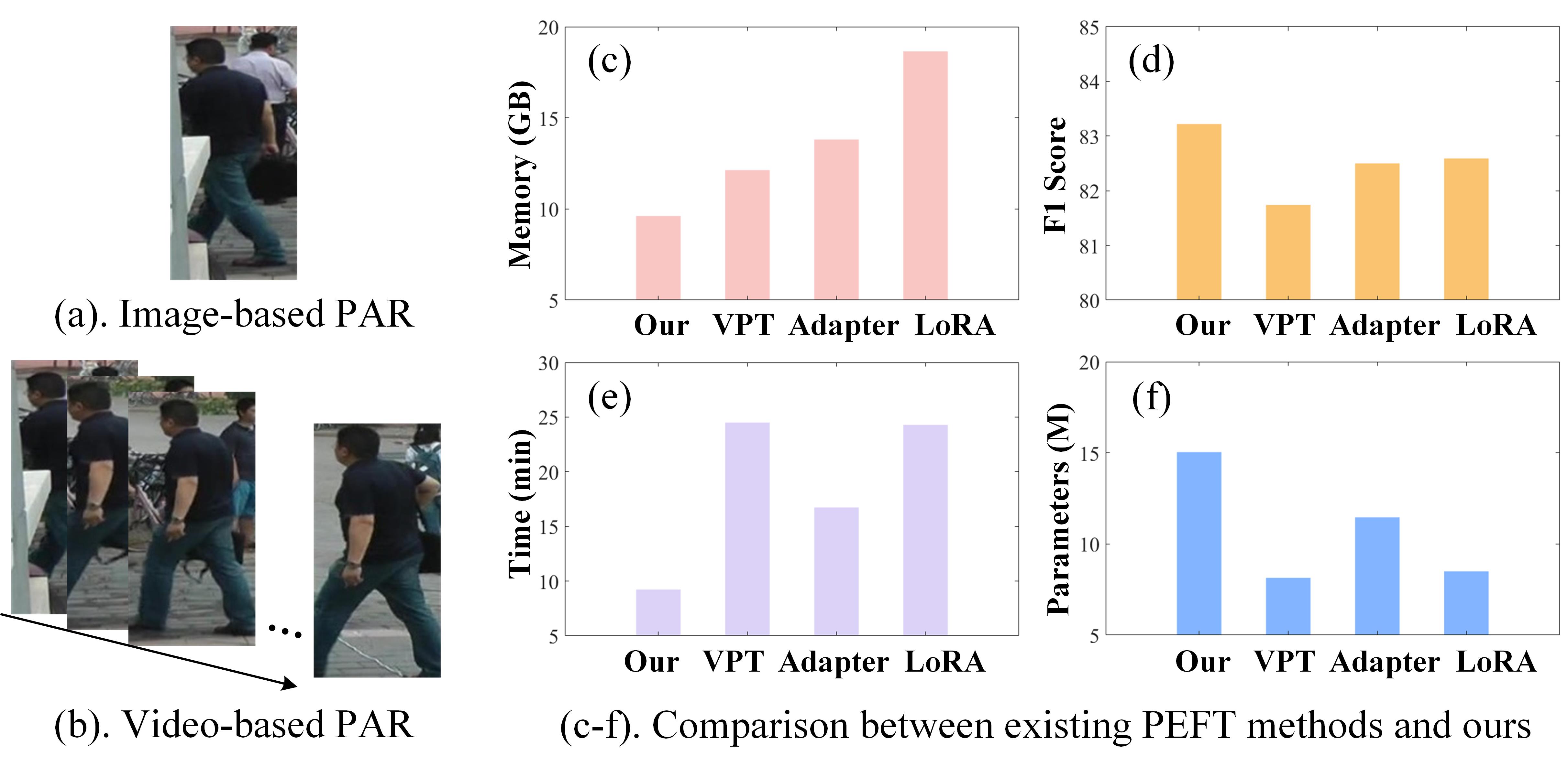}
    \caption{(a, b). Comparison between the RGB frame-based and video-based pedestrian attribute recognition; 
            (c-f). Comparison of memory consumption, F1 score, time cost, and tunable parameters between existing PEFT (parameter-efficient fine-tuning) strategies and our newly proposed spatiotemporal side network tuning method.} 
    \label{fig:enter-label}
\end{figure}

As mentioned in work~\cite{chen2019videoPAR}, the video frames can provide more comprehensive visual information for the specific attribute, but the static image fails to. The authors propose to understand human attributes using video clips and propose large-scale datasets for video-based PAR. They also build a baseline by proposing the multi-task video-based PAR framework based on CNN and temporal attention. Better performance can be obtained on their benchmark datasets, however, we think the following issues still limit their overall results: 
\begin{itemize} 
    \item Existing PAR algorithms adopt CNN as the backbone network to extract the feature representation of input images which learns the local features well. As is known to all, global relations in the pixel-level space are also very important for fine-grained attribute recognition. Some researchers resort to the Transformer network to capture such global information~\cite{DosovitskiyViT, vaswani2017Transformer}, however, their models support the image-based attribute recognition only. 
    
    \item  Chen et al.~\cite{chen2019videoPAR} formulate the video-based pedestrian attribute recognition as a multi-task classification problem and try to learn a mapping from a given video to attributes. The attribute labels are transformed into binary vectors for network optimization. However, the high-level semantic information is greatly missing which is very important for pedestrian attribute recognition. 
\end{itemize}
Therefore, it is natural to raise the following questions: \emph{how to design a novel video-based pedestrian attribute recognition framework that simultaneously captures the global features of vision data, and aligns the vision and semantic attribute labels well?}

To answer this question, in this paper, we take the video frames and attribute set as the input and formulate the video-based PAR as a multi-modal fusion problem. As shown in Fig.~\ref{fig:framework}, a novel CLIP-guided Visual-Text Fusion Transformer for Video-based PAR is proposed. To be specific, the video frames are transformed into video tokens using a pre-trained CLIP~\cite{radford2021CLIP} which is a multimodal foundation model. The attribute set is transformed into corresponding language descriptions using split, expand, and prompt engineering. Then, the text encoder of CLIP is used for the language embedding. After that, we concatenate the video and text tokens and feed them into a fusion Transformer for multi-modal information interaction which mainly contains layer normalization, multi-head attention, and MLP (Multi-Layer Perceptron). The output will be fed into a classification head for pedestrian attribute recognition.

Different from the standard fully fine-tuning strategy, in this work, we propose a novel spatiotemporal side-tuning strategy to optimize the parameters of our framework. As the pre-trained foundation model contains large number of parameters, adjusting all parameters incurs high computational cost. In addition, the multimodal features are already well aligned and hasty fine-tuning may disrupt the original feature space. Specifically, we fix all the parameters of the multimodal foundation model and only optimize the light-weight integrated external side network. We consider both spatial and temporal views of the input pedestrian features and achieve parameter-efficient fine-tuning using our spatiotemporal side-tuning strategy. 
Extensive experiments on two large-scale video-based PAR datasets demonstrate that our proposed spatiotemporal side tuning strategy performs better on the GPU memory usage, time cost in the inference phase, and F1 score, compared with existing PEFT methods.

To sum up, the main contributions of this paper can be concluded as following three aspects: 

1). We propose a novel CLIP-guided Visual-Text Fusion Transformer for Video-based Pedestrian Attribute Recognition, which is the first work to address the video-based PAR from the perspective of visual-text fusion. 

2). We introduce the pre-trained big model CLIP as our backbone network, which makes our model robust to the aforementioned challenging factors. A novel spatiotemporal side-tuning strategy is specifically designed to optimize our PAR framework efficiently. 

3). Extensive experiments on two large-scale video-based PAR datasets fully validated the effectiveness of our proposed video PAR framework and the parameter-efficient fine-tuning method.

This paper is an extension of our previous work which was published in CVPR workshop@NFVLR2023
~\footnote{\url{https://nfvlr-workshop.github.io/}}. The key extensions can be summarized as the following two aspects: 
\textbf{1). New Siding Tuning Strategy}: 
In our conference version, we extract the features of pedestrian videos and attribute text using pre-trained big model CLIP~\cite{radford2021CLIP} and directly fuse them using a multi-modal Transformer. In this work, we propose a novel siding tuning strategy to extract better spatiotemporal features of pedestrian video. 
\textbf{2). More Experiments}: 
In this work, we conduct more ablation studies and parameter analyses to better illustrate the effectiveness of our proposed framework. Another benchmark dataset DukeMTMC-VID attribute dataset~\cite{wu2018exploit} is also used for comparison with other state-of-the-art video-based pedestrian attribute recognition algorithms.

\textbf{Organization of this paper:} 
In Section \ref{relatedworks}, we will review the works most related to ours, including Pedestrian Attribute Recognition, Visual-Language Fusion, and Self-attention and Transformers. In Section \ref{approach}, we mainly introduce our framework with a focus on the overview, network architectures, and loss function. Then, we conduct experiments on two widely used video-based pedestrian attribute recognition datasets in Section \ref{experiments}. Finally, in Section \ref{conclusion}, we conclude this paper and propose possible research directions on the video-based PAR.

\section{Related Works} \label{relatedworks}
In this section, we will give a brief introduction to Pedestrian Attribute Recognition, Pre-trained Vision-Language Foundation Models, and Parameter Efficient Fine Tuning Methods. More related works can be found in the following survey~\cite{wang2022PARsurvey, wang2023MMPTMs}~\footnote{\url{github.com/wangxiao5791509/Pedestrian-Attribute-Recognition-Paper-List}}.

\subsection{Pedestrian Attribute Recognition}  

Current pedestrian attribute recognition can be divided into two main streams, i.e., the RGB frame-based~\cite{abdulnabi2015multi, zhang2014panda, wang2016cnn, tian2015pedestrian, li2019visual, park2017attribute} and video-based PAR~\cite{chen2019videoPAR, specker2020evaluation}. 
%%%% 
For the RGB frame-based PAR, the early research works mainly analyzed pedestrian attributes from the perspective of multi-label classification~\cite{abdulnabi2015multi, zhang2014panda} using convolutional neural networks (CNN). Specifically, Abdulnabi et al.~\cite{abdulnabi2015multi} propose a multi-task learning approach, utilizing multiple CNNs to learn attribute-specific features while sharing knowledge among them. Zhang et al. introduce the PANDA~\cite{zhang2014panda}, a strategy that integrates a part-aware model with human attribute classification based on CNN. This framework accelerates the training of CNN, enabling it to learn robust normalized features even from smaller datasets. 
Due to the RNN (Recurrent Neural Networks) effectively modeling the sequential dependencies between human attributes, some researchers exploit the application of RNN on the PAR task. For example, Wang et al.~\cite{wang2016cnn}  utilize Long Short-Term Memory (LSTM) to establish robust semantic dependencies among labels in pedestrian attribute recognition. By integrating previously predicted labels, the visual features can dynamically adapt to subsequent ones. Zhao et al. propose the GRL~\cite{tian2015pedestrian} to exploit the potential dependencies between pedestrian attributes by considering intra-group attribute mutual exclusion and inter-group attribute association. 
The graph neural network (GNN) is also introduced into the PAR task to model the semantic relations of different attributes. Specifically, VC-GCN~\cite{li2019visual} and A-AOG~\cite{park2017attribute} represent attribute correlations through conditional random fields and graphical models. Li et al.~\cite{li2019visual} take pedestrian attribute recognition as an attribute sequence prediction problem, which utilizes GNN as a basic layer for the whole framework to model the spatial and semantic relations between pedestrian attributes.

Recently, the Transformer whose core operation is the self-attention mechanism, has drawn more and more attention in the artificial intelligence community. Many PAR works are also developed based on the Transformer network, for example, Fan et al.~\cite{fan2023parformer} introduce a PARformer to extract features instead of CNN, which combines global and local perspectives. VTB~\cite{cheng2022VTB} proposes a novel baseline that treats pedestrian attribute recognition by introducing an additional text encoder, which interacts information respectively.

For the video-based PAR, it is a relatively new research topic compared with image-based methods. 
To be specific, Chen et al.~\cite{chen2019videoPAR} introduce a novel multi-task model that includes an attention module to pay attention to each frame for each attribute. 
Specker et al.~\cite{specker2020evaluation} introduce different information from different frames through global features before temporal pooling. 
Lee et al.~\cite{lee2021robust} achieve robust pedestrian attribute recognition by selecting unobstructed frames through sparsity-based temporal attention module.
Thakare et al.~\cite{thakare2024let} compute the pedestrian features through cross-correlation of attribute prediction of different view of pedestrian images.
Liu et al.~\cite{liu2024pedestrian} capture the correlations among different attributes in both spatio and temporal domains through a novel spatio-temporal saliency module. Note that, to better mine the correlations among these patterns, a spatio-temporal attribute relationship learning module is proposed.
Different from these models, in this work, we formulate the video-based PAR as a video-language fusion problem and propose to fuse the dual modalities using a pre-trained vision-language foundation model. More importantly, we propose a new spatiotemporal side tuning strategy to achieve parameter-efficient fine-tuning for the large model-based video pedestrian attribute recognition.

\subsection{Pretrained Vision-Language Foundation Models}  

Inspired by the success of pre-trained large language models (for example, the BERT ~\cite{devlin2018bert}, GPT series ~\cite{radford2019language, brown2020language}, LLaMA~\cite{touvron2023llama}, LaMDA~\cite{thoppilan2022lamda}, Baichuan-2~\cite{yang2023baichuan}) and multi-modal models (e.g., CLIP~\cite{radford2021CLIP}, ALIGN~\cite{jia2021scaling}, LXMERT~\cite{tan2019lxmert}, ViLBERT~\cite{lu2019vilbert}, SAM~\cite{kirillov2023SAM}), many researchers resort to these big models to improve their performance further. Specifically, the CLIP model~\cite{radford2021learning} is pre-trained on 400 million image-text pairs and aligns the dual modalities in the feature space well. SAM~\cite{kirillov2023SAM} is proposed to address the segmentation problem which takes the image and prompt information (e.g., point, bounding box, text) from humans to achieve accurate and interactive segmentation. 
%%%% 
Due to the good generalization performance on the downstream tasks, 
Wang et al. propose the PromptPAR~\cite{wang2023PromptPAR} which adopts the CLIP as the backbone and optimizes its parameters using prompt tuning. Jin et al.~\cite{jin2023sequencepar} formulate the attribute recognition as a phrase generation problem and take the pre-trained CLIP model as the visual and text encoder for the input encoding. 
%%%% 
Some researchers also exploit the human-centric pre-training and validate their model on the pedestrian attribute recognition task, such as Hulk~\cite{wang2023hulk}, HAP~\cite{yuan2024hap}, PLIP~\cite{zuo2023plip}. 
Although these works perform well on the frame-based PAR, however, their model can't process the video-based PAR task. 
%%%%
In our conference version of this paper~\cite{zhu2023videoCLIPPAR}, we also exploit the pre-trained CLIP model to learn the dual modalities to better align the human vision features and semantic features. 
%%%% 
To further improve the training and inference efficiency, we fix the pre-trained CLIP model and further introduce a novel spatiotemporal side network to achieve efficient parameter tuning.

\begin{figure*}
\centering
\includegraphics[width=1\linewidth]{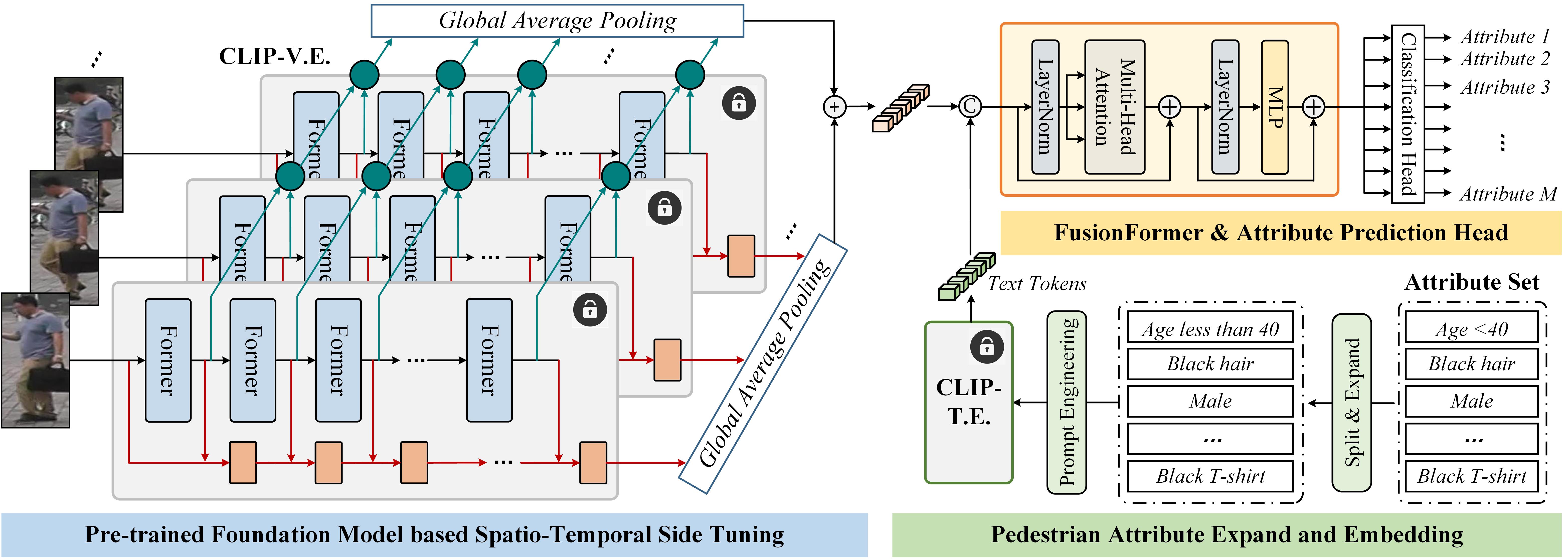}
\caption{ An illustration of our proposed video-based pedestrian attribute recognition framework, termed VTFPAR++. It formulates video-based attribute recognition as a video-language fusion problem, which takes the pedestrian video and attribute set as the input. The pre-trained multi-modal foundation model CLIP is adopted as the basic feature extraction network. We further propose the lightweight spatiotemporal side network to aggregate the features from different Transformer layers and video frames. These features are fused into a unified representation via global average pooling operators. We process the given attributes into language descriptions via split, expand, and prompt engineering, and extract its features using CLIP text encoder. Then, we align the vision-language features using a fusion Transformer and classify the attributes via the attribute prediction head. Our framework requires lower GPU memory consumption, fewer parameter adjustments, and more efficient model training and deployment, yet still achieves leading attribute recognition accuracy on two public datasets.
} 
\label{fig:framework}
\end{figure*}
% https://cn.overleaf.com/project/639c7777b113552694e2a6c4

\subsection{Parameter Efficient Fine Tuning} 
As the size of the model continues to grow rapidly, the question of how to effectively fine-tune the parameters in the model becomes common. Parameter Efficient Fine Tuning (PEFT), aims to achieve comparable performance to full parameter tuning by fine-tuning only a small number of parameters when using a large pre-trained model.
% prompt tuning, 
Previously Prompt Tuning has been widely used in natural language processing, e.g., GPT3~\cite{brown2020language}, BERT~\cite{devlin2018bert}, etc., for constructing templates to convert downstream classification or generative tasks into masked language modeling tasks during pre-training, in order to reduce the gap between pre-training and fine-tuning. Later Prompt Tuning took more forms, such as In-Context Learning~\cite{dong2022survey}, Instruction-tuning~\cite{zhang2023instruction}, and Chain-of-Thought~\cite{wei2022chain}, and was widely used in fine-tuning and reasoning for large-scale models. 
Jia et al.\cite{2022vpt} first proposed Visual Prompt Tuning (VPT) to be introduced into the visual domain to fine-tune the frozen backbone by means of a set of successive learnable vectors. 
Zhou et al.~\cite{zhou2022coop, zhou2022cocoop}, for the first time, applied Prompt to vision, proposed to use continuous vectors as templates for categories and proposed to use a meta-net to extract instance information in conjunction with Prompt, which dramatically improves CLIP~\cite{radford2021CLIP} to migrate to downstream tasks and generalize to invisible categories.
% adapter, 
Adapter\cite{rebuffi2017learning} was first introduced into the Transformer~\cite{vaswani2017Transformer} architecture by Houlsby\cite{houlsby2019parameter}, using the under-projection-activation-up-projection structure embedded in the Transformer layer to fine-tune the whole model.
Gao et al.\cite{gao2024clipadapterc} proposed CLIP-Adapter, which mixes the original CLIP knowledge and the knowledge of the few-shot through two linear layers to achieve the outstanding few-shot recognition performance. 
% LoRA
LoRA~\cite{hu2021lora} divides the fine-tuned pre-trained model into the original weights and the updated weights, and the updated weights are not full-ranked, which can be changed into two low-dimensional matrices A and B by low-rank decomposition, and reduce the number of parameters during fine-tuning by connecting the partial weights to the adjoint matrices formed by A and B.
Dou et al.~\cite{dou2023loramoe} solved the problem of catastrophic forgetting by combining LoRA with MoE~\cite{jacobs1991adaptive}, which can dynamically generate LoRA based on inputs from different tasks.
% side tuning, 
Zhang et al.~\cite{zhang2020side} added a lightweight side net to the original backbone to fuse with the original output to achieve a fine-tuning effect. 
Sung et al.~\cite{sung2022lst} argued that the backpropagation of previous PEFT methods all need to go through the backbone, which cannot achieve training efficiency, so they introduced an additional lightweight ViT~\cite{DosovitskiyViT} next to the backbone, which drastically improves the training efficiency.
%SAN 、 HST
We consider that the video-based pedestrian attribute recognition task requires both fine-grained spatial features and modeling of temporal information over multiple frames, thus, we propose spatial and temporal side networks to augment CLIP spatial information and empower CLIP temporal modeling, respectively.

% spatio-temporal side tuning the pre-trained foundation models. 

\section{Our Proposed Approach} \label{approach}
In this section, we will first give an overview of our proposed framework VTFPAR++, then, we will give more details about this network architecture, including input encoding, spatiotemporal side tuning, video-text fusion Transformer, and attribute prediction head. After that, we will introduce the loss function which is used in the training phase.

\subsection{Overview} 
As shown in Fig.~\ref{fig:framework}, we formulate the video-based pedestrian attribute recognition as a vision-language fusion problem and propose a novel video-based PAR framework, termed VTFPAR++. Given the input pedestrian video, we first adopt the pre-trained CLIP vision encoder with spatiotemporal side tuning networks to extract the visual features. The spatial and temporal side networks will efficiently extract and aggregate the features obtained from different frames and various layers from the CLIP vision encoder. The spatial and temporal features are obtained via the global average pooling operators and concatenated as the final visual representation of the input video. To help our video PAR model understand the pedestrian attributes, in this work, we first split the attributes into discrete word combinations and expand each attribute into a natural language description via prompt engineering. Then, we adopt the CLIP text encoder to obtain the language representations of the expanded attribute phases. The visual and text features are concatenated as unified tokens and fed into a multi-modal Transformer for fusion. The enhanced features will be further fed into the attribute prediction head for final recognition. More details will be introduced in the following sub-sections.

\subsection{Network Architecture} 

Our proposed VTFPAR++ contains four main modules, including the CLIP text encoder, CLIP vision encoder with spatiotemporal side tuning network, multi-modal fusion Transformer, and attribute prediction head.

% \noindent 
\textbf{Input Encoding.~} 
Give a sequence of pedestrian frames $\mathcal{V} \in \mathbb{R}^{T\times H\times W \times C}$, $T, H, W, C$ denotes the number of video frames, height, width, and channel of video frame, respectively, and the attribute set $\mathcal{A} = \{a_1, a_2, a_3, ..., a_M\}$, $M$ is the number of human attributes that need to predict, we will adopt the pre-trained vision-language foundation model CLIP to extract the multi-modal features. 
%%%% 
For the video input, we first divide each frame into non-overlapping patches and feed them into a linear projection layer to get the visual tokens, for example, the tokens of $k$-th video frame can be denoted as $F^{k} \in \mathbb{R}^{N \times D}$. 
Then, we fed these video frames patch tokens into the Transformer layers of the CLIP foundation model after concatenate with the classification token.
% These tokens will be fed into the Transformer layers of the CLIP foundation model. 
Thus, the visual tokens of the whole video at i-th layer of the CLIP can be represented as $\mathcal{F}_v^i = \{F_{1}^i, F_{2}^i, F_{3}^i, ... , F_{T}^i\}$.

The key operation of the Transformer layer used in the pre-trained CLIP model is the \textit{multi-head self-attention}. The calculation of self-attention in each head can be formulated as: 
\begin{equation}
\label{selfattention} 
SelfAtten(Q, K, V) = Softmax(\frac{QK^T}{\sqrt{d}}) V, 
\end{equation}
where the input feature vectors $Q, K, V$ are query, key, and value features, respectively. These three features are transformed from a single input feature using a projection layer. $d$ is the dimension of the input feature vectors. The framework uses similar operations to process both visual and text tokens. Further details will not be elaborated upon.

For the attribute set $\mathcal{A} = \{a_1, a_2, a_3, ..., a_M\}$, we also input them into our framework to help the PAR model better understand the attributes it needs to recognize. Specifically, we first split the attribute into phrases, e.g., "Age < 40" is split and expanded into "Age less than 40". Then, this phrase is transformed into a sentence using prompt engineering. Usually, the prompt is a sentence like "The attribute \underline{~~~~} of this pedestrian is \underline{~~~~~~~~}". By combining the attribute and prompt, we can transform the word/phrase into a language description, such as "The attribute \underline{~age~} of this pedestrian is \underline{~less than 40~}". Then, the pre-trained CLIP text encoder is utilized to extract the semantic attribute representation $\mathcal{F}_a$.

\begin{figure} 
    \centering
    \includegraphics[width=1\linewidth]{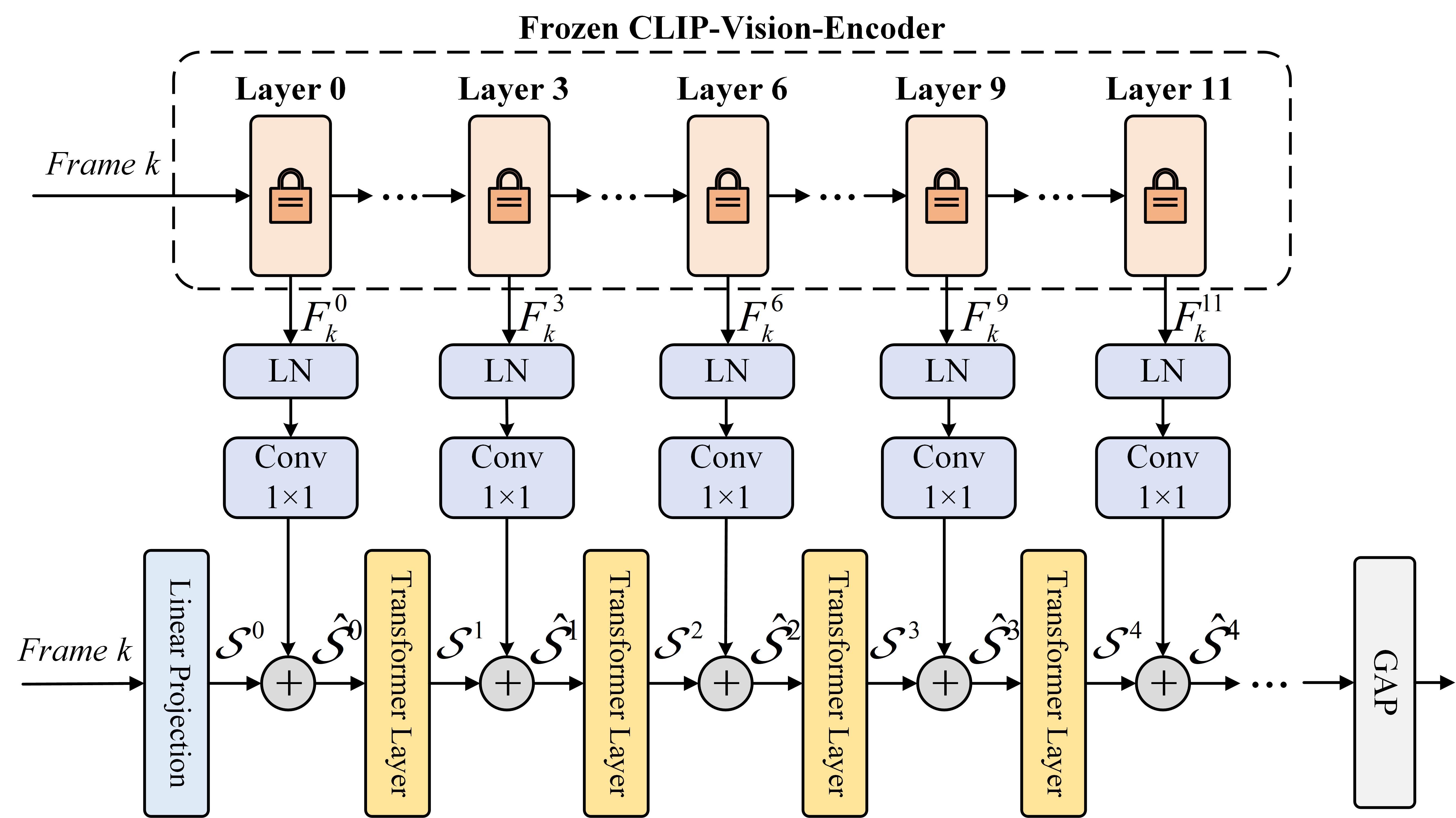}
    \caption{An illustration of our proposed Spatial Side Network(SSN), SSN models the spatial relationship of multi-level CLIP visual features for each frame separately, and finally the modeling results of multiple frames are interacted with text features after GAP aggregation.}
    \label{fig:spatioSideNet}
\end{figure}

% \noindent 
\textbf{Spatio-Temporal Side Tuning.~} 
After we embed the input video and text using the pre-trained CLIP model, we can directly train these networks in a fine-tuning way as we do in our conference paper~\cite{zhu2023videoCLIPPAR}. However, the feature space of the pre-trained large CLIP model has been aligned with a large-scale image-text dataset. If fine-tuning is done hastily, it may disrupt the original feature space. Additionally, the model has a large number of parameters, so direct fine-tuning can cause a sudden increase in computational overhead. Wang et al. propose the PromptPAR~\cite{wang2023PromptPAR} which resorts to prompt tuning for RGB frame-based PAR, however, their method still consumes a large amount of GPU memory due to the prompt tuning need back forward through their framework.

In this paper, we attempt to propose a novel spatiotemporal side tuning strategy to efficiently adjust embedded lightweight networks for video-based PAR. Specifically, the parameters of pre-trained CLIP vision and text encoders are all fixed. We introduce lightweight spatiotemporal side networks along different layers of CLIP vision encoder and different video frames to extract high-quality multi-scale and temporal feature representations, as illustrated in Fig.~\ref{fig:framework}.

\begin{figure} 
    \centering
    \includegraphics[width=0.9\linewidth]{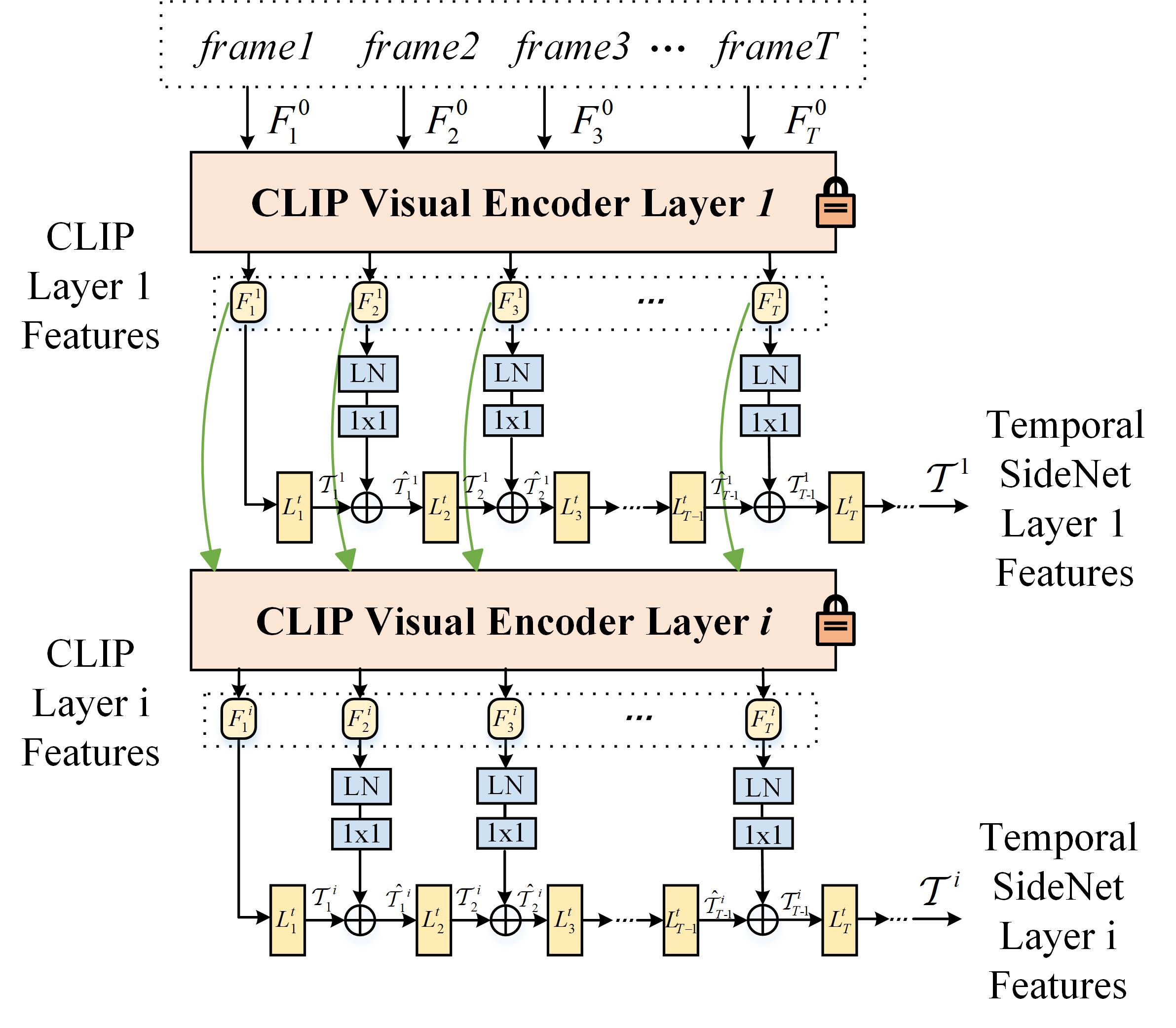}
    \caption{An illustration of our proposed Temporal Side Network(TSN). TSN primarily models temporal relationships of the same layer of CLIP visual features over multiple frames to mitigate the effects of challenges such as occlusion and blurring.} 
    \label{fig:tempSideNet}
\end{figure}

For the spatial side tuning network, as shown in Fig.~\ref{fig:spatioSideNet}, 8 Transformer layers are utilized. In this sub-network, we aggregate the features from the Transformer layers of CLIP vision encoder and the spatial side network via the fusion module (layer normalization layer, convolutional layer with kernel size $1 \times 1$). The two features are then added and fed into the subsequent Transformer layer of the spatial side network, the processing of the \textit{k}-th frame can be formulated as: 
\begin{equation}
\label{spatio_fuse_modules} 
{\hat{\mathcal{S}}^{j}} = LN(F^i_k) * \omega_s + \mathcal{S}^j
\end{equation} 
where $LN(\cdot)$ is layer normalization, and $i$ is the layer we choose to fuse, $j$ is the spatial side network layer corresponding to $i$-th CLIP layer. $\mathcal{S}^j$ is the feature propagated through the $j$-th layer of spatial side network. $\omega_s$ denotes the learnable weights of the $1 \times 1$ convolution. 
Similar operations are executed for feature fusion of all other different Transformer layers.

For the temporal side tuning network, as illustrated in Fig.~\ref{fig:tempSideNet}, after embedding the $T$ video frames into corresponding visual tokens, we first adopt the $LN$ layer and $1 \times 1$ convolutional layer to process the CLIP vision feature. Meanwhile, we feed the embedded visual tokens into temporal side networks and get the output features $\mathcal{T}$. 
\begin{align}
\label{temporal_forward_formulat} 
&{\hat{\mathcal{T}}_{k}^{i}}= LN(F_{k}^{i}) * \omega_t + \mathcal{T}_{k}^{i} \\ 
&\mathcal{T}_{k+1}^{i} = L^t_k({\hat{\mathcal{T}}_{k}^{i}}) 
\end{align} 
where $k$ denotes the index of frames, and $\omega_t$ is the linear layer to transform the dimension of \textit{i}-th CLIP visual feature to match the temporal feature. 
Then, we combine the processed feature from the CLIP vision encoder and our newly proposed temporal side networks in a frame-by-frame manner. Similar operations are also conducted for other CLIP Transformer layers. 
%%%%% 
Through the spatiotemporal side networks, we get the features of temporal and spatial interaction $\mathcal{S}= \{\mathcal{S}^1,\mathcal{S}^2,\mathcal{S}^3, \dots, \mathcal{S}^{T}\}$ and $\mathcal{T} = \{\mathcal{T}^0,\mathcal{T}^3,\mathcal{T}^6, \mathcal{T}^9, \mathcal{T}^{11}\}$. Then the spatial and temporal features will be added after global average pooling and concatenated with the text features to be fed into the multi-modal Transformer.

Our proposed spatiotemporal side network has a similar architecture to the standard CLIP~\cite{radford2021CLIP} visual backbone ViT-B/16~\cite{DosovitskiyViT}, but is more lightweight. We set the width (i.e., the dimension of feature vector) of the side network to 240, the number of attention heads to 6, the depth to 8, and the patch size to 16. Extensive experiments demonstrate that our proposed side networks can aggregate the spatiotemporal visual features in a more efficient and accurate way.

% \noindent 
\textbf{Video-Text Fusion Transformer.~} 
After we get the enhanced spatial and temporal features, we concatenate and feed them into a multi-modal Transformer layer for video-text fusion. A standard Transformer layer is adopted for this module, which contains the layer normalization, multi-head self-attention, and MLP (Multi-Layer Perceptron) layer. The output features will be later fed into the attribute recognition head.

\textbf{Attribute Prediction Head.~}
After we get the output $\mathcal{F}_f$ from the multi-modal Transformer, we utilize the vision tokens for final attribute prediction. A prediction head is proposed to achieve this target which contains $k$ dense layers. The results are further processed by batch normalization $BN$ and Sigmoid function, i.e., 
\begin{equation}
\label{temporal_forward} 
\mathcal{P} = \sigma(BN(Dense(\mathcal{F}_f))) 
\end{equation}  
where $\sigma$ is Sigmoid function, and $Dense$ denotes the dense layers (also termed fully connected layers).

\subsection{Loss Function}  
To measure the distance between the predicted results $\mathcal{P} = \{p_1, p_2, p_3, ..., p_Z\}$ from the attribute prediction head and the ground truth $\mathcal{Y} = \{y_1, y_2, y_3, ..., y_H\}$, we use the weighted cross-entropy loss function which can be formulated as:
\begin{equation}
\label{cross_entropy_loss_function} 
\mathcal{L} = -\frac{1}{M}\sum_{i=1}^N\sum_{j=1}^M w_j(y_{ij}\log(p_{ij})+(1-y_{ij})\log(1-p_{ij})) 
\end{equation} 
where $\omega_j$ considers the imbalanced distributions of each attribute, $p_{ij}$ is the attribute predicted by our model, $N, M$ denotes the number of tracklets, attributes, respectively. 
\begin{equation}
\label{cross_entropy_loss_function_omega} 
\omega_{j}=\left\{\begin{array}{ll}
e^{1-r_{j}}, & y_{i j}=1 \\
e^{r_{j}}, & y_{i j}=0
\end{array}\right.
\end{equation}
where $r_j$ is the ratio of positive samples of $j$-th attribute in the train set.

\section{Experiments} \label{experiments}
In this section, we will first introduce the datasets and evaluation metrics that we use in subsection \ref{dataset}. The implement details will be given in subsection \ref{details}. After that, we compare our method with other state-of-the-art algorithms in subsection \ref{comparison}. Then, we conduct some extended experiments on our newly proposed framework in subsection \ref{ablation}. We also provide a detailed analysis of the parameter in subsection \ref{parameter}. The visualization will also be provided in subsection \ref{visualization} to help the readers better understand our work. We discuss the failed cases and limitation analysis in the subsection \ref{failed}.

\begin{table*} 
\center
\caption{Results on MARS-Attribute and DukeMTMC-VID-Attribute video-based PAR dataset.}  
\label{resultsMARSDukeMTMC} 
\begin{tabular}{l|c|cccc|cccc}
\hline \toprule [0.5 pt] 
\multicolumn{1}{c|}{\multirow{2}{*}{\textbf{Methods}}} & \multicolumn{1}{c|}{\multirow{2}{*}{\textbf{Backbone}}} & \multicolumn{4}{c|}{\textbf{MARS-Attribute Dataset}} & \multicolumn{4}{c}{\textbf{DukeMTMC-VID-Attribute Dataset}}  \\ \cline{3-10} 
\multicolumn{1}{c|}{} &
\multicolumn{1}{c|}{} &
\multicolumn{1}{c}{\textbf{Accuracy}} &
\multicolumn{1}{c}{\textbf{Precision}} &
\multicolumn{1}{c}{\textbf{Recall}} &
\multicolumn{1}{c|}{\textbf{F1 score}} &
\multicolumn{1}{c}{\textbf{Accuracy}} &
\multicolumn{1}{c}{\textbf{Precision}} &
\multicolumn{1}{c}{\textbf{Recall}} &
\multicolumn{1}{c}{\textbf{F1 score}} \\  
\hline
3DCNN~\cite{ji20123d}  & - & 81.95 & - & - & 61.87 & 84.24  & - & - & 62.93  \\
CNN-RNN~\cite{mclaughlin2016recurrent} & - & 86.35 & - & - & 70.42 & 88.84 & - & - & 71.63 \\
ALM~\cite{tang2019improving} & BN-Inception & 86.56 & - & - & 68.89 & 88.13 & - & - & 69.66 \\
SSC$_{soft}$~\cite{jia2021spatial} & ResNet50 & 86.00 & - & - & 68.15 & 87.52 & - & - & 68.71 \\
TA(Image)~\cite{chen2019videoPAR} & ResNet50 & 85.85 & - & - & 67.28 & 87.77 & - & - &68.70  \\
TA(Video)~\cite{chen2019videoPAR} & ResNet50 & 87.01 & - & - & 72.04 & 89.31 & - & - & 73.24  \\
Lee et al.~\cite{lee2021robust} & ResNet50 & 86.75 & - & - & 70.42 & 88.98 & - & - & 72.30 \\
TRA~\cite{zhao2023tra} & ResNet50 & 87.05 & - & - & 71.92 & 89.32 & - & -  & 75.01 \\
VTB~\cite{cheng2022VTB} & ViT-B/16 & 90.37 & 78.96 & 78.42 & 78.32 & 90.29 &73.38 &76.99  &74.81   \\
\hline 
VTF~\cite{zhu2023videoCLIPPAR} & ViT-B/16 &  \textcolor{SlateBlue}{\textbf{92.47}} & \textcolor{SlateBlue}{\textbf{81.76}} & \textcolor{SlateBlue}{\textbf{82.95}} & \textcolor{SlateBlue}{\textbf{81.94}} &  \textcolor{SlateBlue}{\textbf{92.45}} & \textcolor{SlateBlue}{\textbf{77.23}}  &\textcolor{SlateBlue}{\textbf{81.44}}  & \textcolor{SlateBlue}{\textbf{78.83}} \\
Ours & ViT-B/16 & \textcolor{DarkRed}{\textbf{93.19}} &  \textcolor{DarkRed}{\textbf{82.27}} & \textcolor{DarkRed}{\textbf{84.87}} &  \textcolor{DarkRed}{\textbf{83.22}} & \textcolor{DarkRed}{\textbf{93.31}} & \textcolor{DarkRed}{\textbf{78.19}}  &\textcolor{DarkRed}{\textbf{83.18}}  &\textcolor{DarkRed}{\textbf{80.45}} \\
Improvements & - & \textcolor{SeaGreen4}{\textbf{+0.72}} & \textcolor{SeaGreen4}{\textbf{+0.51}} & \textcolor{SeaGreen4}{\textbf{+1.92}} & \textcolor{SeaGreen4}{\textbf{+1.28}} & \textcolor{SeaGreen4}{\textbf{+0.86}} & \textcolor{SeaGreen4}{\textbf{+0.60}}  & \textcolor{SeaGreen4}{\textbf{+0.13}}  & \textcolor{SeaGreen4}{\textbf{+0.53}}  \\
\hline \toprule [0.5 pt] 
\end{tabular} 
\end{table*}

\begin{table*} 
\center
\caption{Results on MARS video-based PAR dataset. Accuracy and F1-score are reported for all the assessed attributes.}   
\label{F1scoreMARS} 
\begin{tabular}{l|cc|cc|cc|cc|cc|cc|cc}
\hline \toprule [0.5 pt] 
\multicolumn{1}{c|}{\multirow{2}{*}{\textbf{Attribute}}} & \multicolumn{2}{c|}{\textbf{TA(Image)}} & \multicolumn{2}{c|}{\textbf{3DCNN}}  & \multicolumn{2}{c|}{\textbf{CNN-RNN}}  & \multicolumn{2}{c|}{\textbf{TA(Video)}} & \multicolumn{2}{c|}{\textbf{TRA}}   & \multicolumn{2}{c|}{\textbf{VTF}} & \multicolumn{2}{c}{\textbf{VTFPAR++}} \\ \cline{2-15} 
 \multicolumn{1}{c|}{} &
\multicolumn{1}{c}{Acc} &
\multicolumn{1}{c|}{F1} &

\multicolumn{1}{c}{Acc} &
\multicolumn{1}{c|}{F1} &

\multicolumn{1}{c}{Acc} &
\multicolumn{1}{c|}{F1} &

\multicolumn{1}{c}{Acc} &
\multicolumn{1}{c|}{F1} &

\multicolumn{1}{c}{Acc} &
\multicolumn{1}{c|}{F1} &

\multicolumn{1}{c}{Acc} &
\multicolumn{1}{c|}{F1} &

\multicolumn{1}{c}{Acc} &
\multicolumn{1}{c}{F1}  \\ 
\hline

top length      & 94.21  & 58.72 & 93.63  & 56.37 & 94.60  & 65.18 & 94.47  & 71.61 & \textcolor{DarkRed}{\textbf{95.12}}  & 69.09  & \textcolor{SlateBlue}{\textbf{94.62}} & \textcolor{DarkRed}{\textbf{97.26}} & 94.57  & \textcolor{SlateBlue}{\textbf{96.76}} \\
bottom type     & 93.12  & 81.69 & 89.19  & 72.86 & 93.67  & 84.16 & \textcolor{DarkRed}{\textbf{94.60}}  & 86.62 & 93.73 & 85.06  & 92.97  & \textcolor{SlateBlue}{\textbf{97.21}} & \textcolor{SlateBlue}{\textbf{94.29}}  & \textcolor{DarkRed}{\textbf{97.22}} \\

shoulder bag    & 80.39  & 72.57 & 71.82  & 61.30 & 82.70  &  75.89 & \textcolor{SlateBlue}{\textbf{83.48}}  & \textcolor{SlateBlue}{\textbf{76.08}} & 82.90 &  \textcolor{DarkRed}{\textbf{76.83}}& 80.26  & 65.61 & \textcolor{DarkRed}{\textbf{84.12}}  & 70.97 \\
backpack        & 89.37  & 85.95 & 82.60  & 76.58 & 90.18  & \textcolor{SlateBlue}{\textbf{ 87.17}} & 90.59  & \textcolor{DarkRed}{\textbf{87.62}} & \textcolor{SlateBlue}{\textbf{ 90.63}} & 86.90 & \textcolor{DarkRed}{\textbf{91.32}}  & 82.08 & 90.56  & 83.13 \\
hat             & 96.91 & 57.57 & 96.53  & 57.69 & 97.90  & \textcolor{SlateBlue}{\textbf{ 77.74}} & 97.51  & \textcolor{DarkRed}{\textbf{77.84}} & 97.84 & 77.70 & \textcolor{SlateBlue}{\textbf{ 98.32}}  & 72.76 & \textcolor{DarkRed}{\textbf{98.42}}  & 65.66 \\
handbag         & 85.71  & 62.82 & 83.88  & 59.90 & 88.07  & \textcolor{SlateBlue}{\textbf{ 71.68}} & 87.61  & \textcolor{DarkRed}{\textbf{73.55}} & 87.03 & 70.98 & \textcolor{SlateBlue}{\textbf{ 87.63}} & 59.08 & \textcolor{DarkRed}{\textbf{89.64}}  & 66.61\\
hair            & 88.61  & 86.91 & 85.12  & 82.77 & 88.78  & \textcolor{SlateBlue}{\textbf{87.11}} & 89.54  &\textcolor{DarkRed}{\textbf{88.17}} & 89.54 & 88.23 & \textcolor{SlateBlue}{\textbf{90.51}}  & 86.37 & \textcolor{DarkRed}{\textbf{90.56}}  & 86.49\\

gender          & 91.32  & 90.89 & 86.49  & 85.75 & 92.77  & 92.44 & 92.83  & 92.50 & 92.89  & 92.53  & \textcolor{SlateBlue}{\textbf{94.56}}  & \textcolor{SlateBlue}{\textbf{92.88}} & \textcolor{DarkRed}{\textbf{94.87}}  & \textcolor{DarkRed}{\textbf{93.68}} \\
bottom length   & 92.69  & 92.29 & 89.96  & 89.35 & 93.70  & 93.33 & 94.22  & \textcolor{SlateBlue}{\textbf{93.90}} & 93.41 & 93.06 & \textcolor{DarkRed}{\textbf{95.68}}  & 93.69 & \textcolor{SlateBlue}{\textbf{95.21}} & \textcolor{DarkRed}{\textbf{94.99}} \\
pose            & 72.03  & 56.91 & 62.51  & 47.69 & 72.40  & 58.36 & 73.65  & 61.36 & 72.50 & 56.86 & \textcolor{SlateBlue}{\textbf{90.32}}  & \textcolor{SlateBlue}{\textbf{74.84}} & \textcolor{DarkRed}{\textbf{92.10}}  & \textcolor{DarkRed}{\textbf{76.27}}\\
motion          & 91.08  & 39.39 & 90.34  & 33.64 & 92.12  & 43.92 & 92.12  & 43.69 & 92.62 & 45.27 & \textcolor{SlateBlue}{\textbf{97.33}}  & \textcolor{SlateBlue}{\textbf{93.50}} & \textcolor{DarkRed}{\textbf{98.05}}  & \textcolor{DarkRed}{\textbf{95.03}}\\
top color       & 74.73  & 72.72 & 68.04  & 65.63 & 71.90  & 69.28 & 73.43  & 71.44 & 74.41 & 72.34 & \textcolor{SlateBlue}{\textbf{93.85}}  & \textcolor{SlateBlue}{\textbf{74.97}} & \textcolor{DarkRed}{\textbf{94.67}}  & \textcolor{DarkRed}{\textbf{78.08}} \\
bottom color    & 68.27  & 44.63 & 65.44  & 40.39 & 65.77  & 39.68 & 69.45  & 43.98 & 71.10 & 48.46 & \textcolor{SlateBlue}{\textbf{93.66}}  & \textcolor{SlateBlue}{\textbf{69.76}} &  \textcolor{DarkRed}{\textbf{94.38}}  &  \textcolor{DarkRed}{\textbf{74.01}}\\
age             & 83.44  & 38.87 & 81.70  & 36.22 & 84.28  & 39.93 & 84.71  & 40.21 & 84.92 & 43.53 & \textcolor{DarkRed}{\textbf{93.50}} & \textcolor{SlateBlue}{\textbf{87.07}} & \textcolor{SlateBlue}{\textbf{93.17}}  & \textcolor{DarkRed}{\textbf{86.09}} \\
\hline
\textbf{Average} & 85.85  & 67.28 & 81.95  & 61.87 & 86.35  & 70.42 & 87.01  & 72.04 & 87.05 & 71.92 & \textcolor{SlateBlue}{\textbf{92.47}} & \textcolor{SlateBlue}{\textbf{81.94}}  & \textcolor{DarkRed}{\textbf{93.19}}  & \textcolor{DarkRed}{\textbf{83.22}}  \\
\hline \toprule [0.5 pt] 
\end{tabular} 
\end{table*}

\begin{table*} 
\center
\caption{Results on DUKE video-based PAR dataset.Accuracy and F1-score are reported for all the assessed attributes.}   
\label{F1scoreDUKE} 
\begin{tabular}{l|cc|cc|cc|cc|cc|cc|cc}
\hline \toprule [0.5 pt] 
\multicolumn{1}{c|}{\multirow{2}{*}{\textbf{Attribute}}} & \multicolumn{2}{c|}{\textbf{TA(Image)}} & \multicolumn{2}{c|}{\textbf{3DCNN}}  & \multicolumn{2}{c|}{\textbf{CNN-RNN}}  & \multicolumn{2}{c|}{\textbf{TA(Video)}} & \multicolumn{2}{c|}{\textbf{TRA}}   & \multicolumn{2}{c|}{\textbf{VTF}} & \multicolumn{2}{c}{\textbf{VTFPAR++}} \\ \cline{2-15} 
 \multicolumn{1}{c|}{} &
\multicolumn{1}{c}{Acc} &
\multicolumn{1}{c|}{F1} &

\multicolumn{1}{c}{Acc} &
\multicolumn{1}{c|}{F1} &

\multicolumn{1}{c}{Acc} &
\multicolumn{1}{c|}{F1} &

\multicolumn{1}{c}{Acc} &
\multicolumn{1}{c|}{F1} &

\multicolumn{1}{c}{Acc} &
\multicolumn{1}{c|}{F1} &

\multicolumn{1}{c}{Acc} &
\multicolumn{1}{c|}{F1} &

\multicolumn{1}{c}{Acc} &
\multicolumn{1}{c}{F1}  \\ 
\hline
motion          & 97.65  & 19.76 & 97.68  & 21.37 & 97.76  & 26.65 & 97.65  & 27.68 & 98.14 & 40.93 & \textcolor{DarkRed}{\textbf{99.16}}  & \textcolor{DarkRed}{\textbf{97.91}} & \textcolor{SlateBlue}{\textbf{99.09}}  & \textcolor{SlateBlue}{\textbf{97.74}}\\
pose            & 72.46  & 62.63 & 69.46  & 59.95 & 74.36  & 66.26 & 75.31  & 67.73 & 75.66 & 65.28 & \textcolor{SlateBlue}{\textbf{92.51}}  & \textcolor{SlateBlue}{\textbf{79.22}} & \textcolor{DarkRed}{\textbf{93.19}}  & \textcolor{DarkRed}{\textbf{79.92}}\\
backpack        & 87.41  & 86.12 & 81.05  & 77.59 & 89.78  &  87.95 & \textcolor{SlateBlue}{\textbf{90.05}}  & 88.37 & \textcolor{DarkRed}{\textbf{90.28}} & 88.58& 86.53  & \textcolor{SlateBlue}{\textbf{90.55}} & 89.41  & \textcolor{DarkRed}{\textbf{92.20}} \\
shoulder bag    & 86.15  & \textcolor{SlateBlue}{\textbf{77.90}} & 83.14  & 64.28 & 87.35 &  75.33 & \textcolor{SlateBlue}{\textbf{87.88}}  & 76.47 & \textcolor{DarkRed}{\textbf{88.38}} & \textcolor{DarkRed}{\textbf{79.60}} & 85.20  & 63.27 & 87.32  & 67.50 \\
handbag         & \textcolor{SlateBlue}{\textbf{94.95}}  & 56.34 & 94.34  & 51.09 & 94.34  &  57.82 & 94.42  & \textcolor{SlateBlue}{\textbf{64.67}} & 94.76 & \textcolor{DarkRed}{\textbf{69.14}} & 93.43  & 44.37 & \textcolor{DarkRed}{\textbf{95.82}}  & 61.26\\
boots           & 94.12  & 92.57 & 83.59 & 78.97 & \textcolor{SlateBlue}{\textbf{94.72}}  & \textcolor{SlateBlue}{\textbf{93.25}} & \textcolor{DarkRed}{\textbf{94.95}}  & \textcolor{DarkRed}{\textbf{93.52}} & 94.57 & 93.06& 93.39  & 88.04 & 94.42 & 89.55 \\
gender          & 89.57  & 89.49 & 82.49  & 82.47 & 90.35  & 90.28 & 90.85  & 90.78 & 90.51  & 90.42  & \textcolor{DarkRed}{\textbf{93.05}}  & \textcolor{DarkRed}{\textbf{92.45}} & \textcolor{SlateBlue}{\textbf{92.79}}  & \textcolor{SlateBlue}{\textbf{92.10}} \\
hat             & 93.02  & 88.26 & 87.54  & 76.12 & 93.32  &  88.45 & 93.73  & \textcolor{SlateBlue}{\textbf{89.41}} & 93.92 & \textcolor{DarkRed}{\textbf{89.67}} & \textcolor{SlateBlue}{\textbf{93.70}} & 83.72 & \textcolor{DarkRed}{\textbf{94.46}}  & 85.37 \\
shoes color     & \textcolor{DarkRed}{\textbf{93.32}}  & 83.35 & 88.07  & 69.76 & 93.05  & \textcolor{SlateBlue}{\textbf{84.65}} & \textcolor{SlateBlue}{\textbf{93.13}}  & \textcolor{DarkRed}{\textbf{85.18}} & 92.67 & 84.20 & 92.07  & 68.09 & 92.03  & 72.44 \\
top length      & 91.54  & 78.10 & 89.14  & 69.28 & 92.25  & 80.06 & \textcolor{SlateBlue}{\textbf{92.52}}& \textcolor{DarkRed}{\textbf{81.06}} & \textcolor{DarkRed}{\textbf{92.59}}  & \textcolor{SlateBlue}{\textbf{80.85}}  & 91.27  & 65.36 & 90.25  & 65.12 \\
bottom color    & 76.82  & 47.25 & 75.92  & 48.19 & 78.85  & 51.66 & 79.95  & 55.95 & 80.40 & 57.54 & \textcolor{SlateBlue}{\textbf{93.82}}  & \textcolor{SlateBlue}{\textbf{76.55}} & \textcolor{SlateBlue}{\textbf{94.69}}  &  \textcolor{DarkRed}{\textbf{79.21}}\\
top color       & 76.25  & 42.57 & 78.39  & 56.14 & 79.98  & 57.20 & 81.28  & 58.07 & 79.95 & 60.90 & \textcolor{SlateBlue}{\textbf{95.19}}  & \textcolor{SlateBlue}{\textbf{79.74}} & \textcolor{DarkRed}{\textbf{96.15}}  & \textcolor{DarkRed}{\textbf{82.87}}  \\

\hline
\textbf{Average} & 87.77  & 68.70 & 84.24  & 62.93 & 88.84  & 71.63 & 89.31  & 73.24 & 89.32 & 75.01 & \textcolor{SlateBlue}{\textbf{92.45}} & \textcolor{SlateBlue}{\textbf{77.44}}  & \textcolor{DarkRed}{\textbf{93.31}}  & \textcolor{DarkRed}{\textbf{80.45}}  \\
\hline \toprule [0.5 pt] 
\end{tabular} 
\end{table*}

\subsection{Dataset and Evaluation Metric} \label{dataset}

In our experiments, the \textbf{MARS-Attribute dataset}~\cite{chen2019temporal}  and the \textbf{DukeMTMC-VID-Attribute dataset}~\cite{chen2019temporal} are used~\footnote{\url{https://irip.buaa.edu.cn/mars_duke_attributes/index.html}} which are re-annotated based on MARS~\cite{zheng2016mars} and DukeMTMC-VID~\cite{ristani2016performance} dataset.

$\bullet$ \textbf{MARS-Attribute dataset} is an annotated pedestrian attributes dataset of the MARS~\cite{zheng2016mars} dataset by Chen et al.~\cite{chen2019temporal}. Five sets of multi-label attributes such as action, pedestrian orientation, color of upper/lower body, age, and nine sets of binary attributes such as gender, hair length, and length of upper/lower garment were annotated. We split the multi-label attribute into binary attributes as well, which means we use 43 binary attributes for training and testing. The training subset contains 8,298 sequences from 625 different ID pedestrians and the testing subset contains 8,062 sequences corresponding to 626 pedestrians. For each sequence, there are 60 frames on average, from which we randomly selected 6 frames for training and testing. 

$\bullet$ \textbf{DukeMTMC-VID-Attribute dataset} is also annotated pedestrian attributes dataset of the DukeMTMC-VID~\cite{ristani2016performance} dataset by Chen et al.~\cite{chen2019temporal}. There are four sets of multi-label attributes, such as motion, pedestrian pose, and color of upper/lower body, and eight sets of binary attributes such as backpack, shoes, and boots were annotated. We use 37 binary attributes for training and testing by splitting the multi-label attribute into binary attributes. The training subset contains 702 different ID pedestrians and 16522 images and the testing subset contains 17661 images corresponding to 702 pedestrians. For each sequence, there are 169 frames on average, from which we randomly selected 6 frames for training and testing. 

% the MARS~\cite{chen2019temporal} dataset proposed by Chen et al., and DukeMTMC-VideoReID~\cite{ristani2016performance} dataset proposed by Wu et al. are used for both training and testing. 
% The training subset of MARS dataset contains 8062 tracklets from 625 people, and the testing subset contains tracklets corresponding to 626 pedestrians. For each tracklet, there are 60 frames on average. 
% The training subset of DUKE dataset contains 2196 tracklets from 702 people, and the testing subset contains 2636 tracklets corresponding to 702 pedestrians. Six frames are extracted from each video. 

For the evaluation of our and the compared PAR models, we adopt the widely used Accuracy, Precision, Recall, and F1-score as the evaluation metric. Note that, the results reported in our experiments are obtained by averaging these metrics across multiple attribute groups. The formulation of \textbf{Accuracy, Precision, Recall} and \textbf{F1-score (F1)} can be expressed as:
\begin{equation}
     Accuracy=\dfrac{TP+TN}{TP+TN+FP+FN}
 \end{equation}
 \begin{equation}
     Precision=\dfrac{TP}{TP+FP}, ~~~~ Recall=\dfrac{TP}{TP+FN}
 \end{equation}
\begin{equation}
    F1-score=\dfrac{2\times Precision\times Recall}{Precision+Recall} 
\end{equation}
where $TP$ denotes the number of correctly predicted positive samples, 
      $TN$ is the number of correctly predicted negative samples, 
      $FP$ and $FN$ denote the number of false positive and false negative samples, respectively.

\subsection{Implementation Details} \label{details} 
In our experiments, we adopt the ViT-B/16 version of the pre-trained CLIP foundation model. During the training phase, we fix the parameter of CLIP~\cite{radford2021CLIP} and choose Adam~\cite{kingma2014adam} as our optimizer. The learning rate is set as 0.001, batch size is 16. On the spatial and temporal side network, we set 8 Transformer layers, whose channel dimension is 240 by default. We use 6 attention heads, patch size is $16 \times 16$, and 5 fusion networks. We choose layer 0, layer 3, layer 6, layer 9, and layer 11 features of the CLIP visual backbone for interacting with the side net. 
%%%% 
Our model is implemented based on PyTorch~\cite{paszke2019pytorch} deep learning framework and the experiments are conducted based on a server with NVIDIA RTX 3090 GPUs. More details about our framework can be found in our source code.

\subsection{Comparison with Other SOTA Models} \label{comparison} 
We compare our proposed framework with other strong baseline methods on the MARS-Attribute~\cite{chen2019temporal} and DukeMTMC-VID Attribute~\cite{ristani2016performance} datasets, including 3DCNN~\cite{ji20123d}, CNN-RNN~\cite{mclaughlin2016recurrent}, VideoPAR~\cite{chen2019videoPAR}, VTB~\cite{cheng2022VTB}, and our baseline VTF~\cite{zhu2023videoCLIPPAR}.

\noindent 
$\bullet$ \textbf{Result on MARS-Attribute Dataset~\cite{chen2019videoPAR}. }
As shown in Table~\ref{resultsMARSDukeMTMC}, our baseline method VTF~\cite{zhu2023videoCLIPPAR} obtains 92.47, 81.76, 82.95, 81.94 on Accuracy, Precision, Recall, and F1-score respectively on the MARS dataset which already significantly outperforms previous methods. In contrast, our proposed VTFPAR++ further improves by +0.72, +0.51, +1.92, and +1.28, respectively. Compared to methods based on CNN architectures such as 3DCNN~\cite{ji20123d}, our method has a huge improvement in the F1-score of about 16 points.

As illustrated in Table~\ref{F1scoreMARS}, it describes the comparison of Accuracy and F1 scores with previous methods for each attribute in detail, our proposed method VTFPAR++ can exceed our baseline on most of the attributes, especially on the \textit{pose}, and \textit{motion} attributes. Specifically, it improves these attributes by +1.43, and +1.53, respectively. This improvement is attributed to the advantage of our proposed spatiotemporal side tuning strategy that can better learn spatial and temporal features through the dual branch side network.

\noindent 
$\bullet$ \textbf{Result on DukeMTMC-VID Attribute Dataset~\cite{ristani2016performance}.}  
As the experimental results of DukeMTMC-VID Attribute dataset reported in Table~\ref{resultsMARSDukeMTMC}, our proposed VTFPAR++ achieves 93.31, 78.19, 83.18, and 80.45 on the Accuracy, Precision, Recall, and F1-score metric, respectively. It exceeds the VideoPAR (image) and VideoPAR (video) by +11.75 and +7.21 on the F1-score, respectively. In addition, our VTFPAR++ also beats the baseline VTF~\cite{zhu2023videoCLIPPAR} and VTB~\cite{cheng2022VTB} on this dataset, which fully demonstrates the validity of our approach.

We also list the detail attributes comparison of Accuracy and F1-Score with other previous method on DukeMTMC-VID Attribute dataset in \ref{F1scoreDUKE}. We can easily find that our framework exceed other method on most attributes, which benefit to our branch side net work.

\begin{table} 
\center
\small  
\caption{Ablation study on the MARS-Attribute dataset.  SSN and TSN is short for the spatial side network and temporal side network, respectively. Params refers to the number of fine-tuning parameters of the model
.} \label{AblationStudy} 
\resizebox{\columnwidth}{!}{ 
\begin{tabular}{c|cccc|ccc}
\hline \toprule [0.5 pt]
\multicolumn{1}{c|}{\multirow{1}{*}{NO.}} & 
\multicolumn{1}{c}{FFN} & 
\multicolumn{1}{c}{VTFormer} & 
\multicolumn{1}{c}{SSN} & 
\multicolumn{1}{c|}{TSN} & 
\multicolumn{1}{c}{Acc} &
% \multicolumn{1}{c}{Prec} & 
% \multicolumn{1}{c}{Recall} & 
\multicolumn{1}{c}{F1} &
\multicolumn{1}{c}{Params(M)}
\\ 
\hline 
% 1   &\cmark        &        &        &      & 89.72 & 77.60 & 81.32 & 78.69 & 78.69 \\ 
% 2   &        & \cmark &        &            & 92.47 & 81.76 & 82.95 & 81.94 & 78.69 \\
% 3   &        & \cmark & \cmark &            & 92.90 & 82.31 & 84.18 & 82.84 & 78.69  \\
% 4   &        & \cmark &        & \cmark     & 92.99 & 79.94 & 86.94 & 82.90 & 78.69 \\
% 5   &        & \cmark & \cmark & \cmark     & 93.09 & 82.27 & 84.87 & 83.22 & 78.69  \\

1   &\cmark        &        &        &      & 89.74 & 78.69 & 150.45 \\ 
2   &        & \cmark &        &            & 92.47 & 81.94 & 157.53 \\
3   &        & \cmark & \cmark &            & 92.90 & 82.84 & 7.85  \\
4   &        & \cmark &        & \cmark     & 92.99 & 82.90 & 8.32 \\
5   &        & \cmark & \cmark & \cmark     & 93.09 & 83.22 & 15.04 \\
\hline \toprule [0.5 pt] 
\end{tabular}  } 
\end{table}

\subsection{Ablation Study} \label{ablation} 
In this section, we will evaluate the effectiveness of each component of our proposed framework. To accomplish this, we will first remove the multi-modal Transformer and then remove our proposed spatial side network and temporal side network separately to check its influence on our final performance.

\noindent 
$\bullet$ \textbf{Effects of Video-Text Multi-modal Transformer.} 
As shown in Table~\ref{AblationStudy}, we use a simple linear layer fusion method (first line) to replace the Video-Text Multi-modal Transformer (second line). The multi-modal Transformer can improve +2.73 and +3.25 on the Accuracy and F1-score metric, respectively. Accordingly, it can be observed that the multi-modal Transformer in our proposed framework facilitates modal information interaction between video and text compared to simple modeling approaches, and the model that use VTFormer instead of FFN only increases the learnable parameters by 7.08M.

\noindent 
$\bullet$ \textbf{Effects of Spatial Side Network.} 
As the result in the third line of Table~\ref{AblationStudy}, we add the spatial side network to fine-tune the CLIP~\cite{radford2021CLIP} features extractor for the low-resolution pedestrian image. We can easily find that the Accuracy and F1-score metrics individually improve by +0.43 and +0.90 compared with the baseline without spatial side net. Moreover, the use of the spatial side net tuning strategy instead of full fine-tuning reduces the number of fine-tuning parameters about 100M, which demonstrates that our proposed spatial side net allows CLIP to adapt effectively to pedestrian images.

\noindent 
$\bullet$ \textbf{Effects of Temporal Side Network.} 
As shown in the fourth line of Table~\ref{AblationStudy}, we add the temporal side net in our baseline to make CLIP~\cite{radford2021CLIP} gain the temporal modeling capability. We can find that the overall performance can be improved by +0.52 and +0.96 on Accuracy and F1-score metrics, respectively. And with the use of the temporal side network, only 8.32M learnable parameters the framework need.

\noindent 
$\bullet$ \textbf{Effects of Spatio-Temporal Side Network.} 
In the latest line of Table~\ref{AblationStudy}, we use the complete version of the spatiotemporal side network and tune the parameters of the side network and CLIP model~\cite{radford2021CLIP}. This version improved the two evaluation metrics by +0.62 and +1.28, respectively. Only requires about 10$\%$ of the number of parameters compare with full-tuning, but has a much faster training speed. Our results indicate that having both temporal and spatial modeling capabilities is better than having only one of them. This demonstrates that our proposed spatiotemporal side net effectively enhances CLIP's ability to recognize pedestrian attributes in videos.

\subsection{Parameter Analysis} \label{parameter}
In this section, we will examine the effect of different input frames on our network. We will then provide a comprehensive analysis of the various parameters of the spatio-temporal side net, such as the dimension of the side net, the different layers of interaction with CLIP~\cite{radford2021CLIP}, and the different modules of aggregating side net features. Additionally, we will compare our method to other Parameter Efficient Fine-Tuning (PEFT) strategies to confirm its superiority.

\begin{figure} 
\centering
\small
\includegraphics[width=3in]{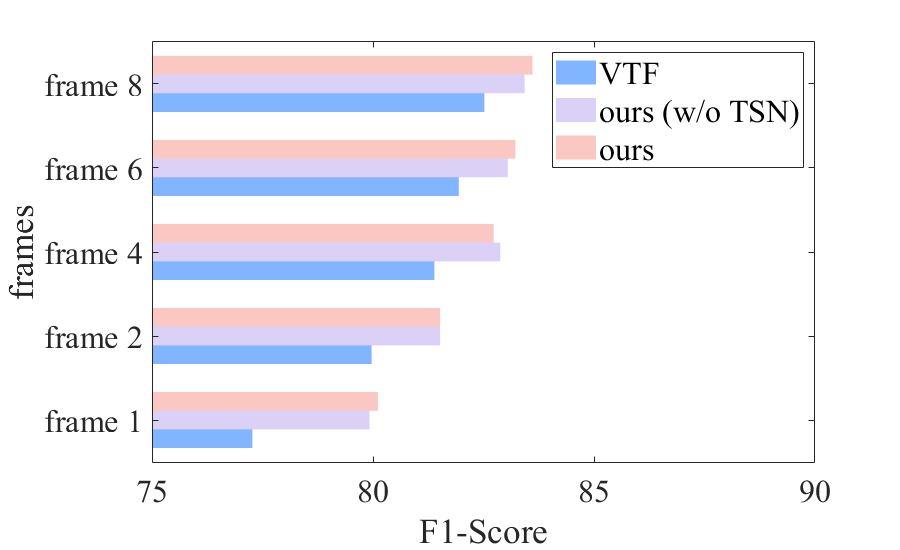}
\caption{Comparison between our method and VTF at different frames on MARS datasets.} 
\label{frame} 
\end{figure}

\noindent 
$\bullet$ \textbf{Influence of different input frames.}  
We performed parametric analysis experiments on different numbers of input frames. As the result shows in Fig.~\ref{frame}, we can intuitively find that the accuracy improves as the number of input frames increases. Furthermore, the inclusion of the temporal side network in our proposed model has led to even greater improvements compared to models without it. This demonstrates the network's capability to effectively model the temporal relationships between different frames. Specifically, when utilizing 6 or 8 input frames, our model achieves F1 scores approximately 0.2 higher than versions lacking the temporal side network. However, considering the training time and memory utilization, we ultimately opt for 6 frames as our input, striking a balance between computational efficiency and model performance. 

\begin{figure}
\centering
\includegraphics[width=1\linewidth]{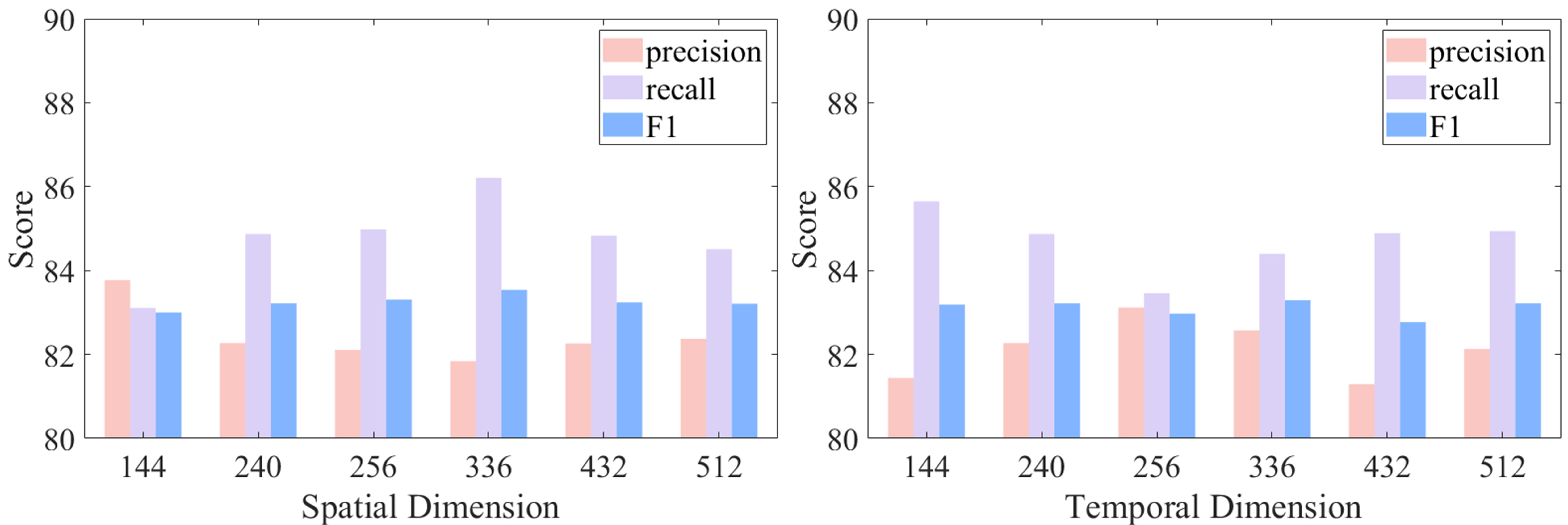}
\caption{Comparison of different side net dimensions on MARS datasets, (a) is the comparison of different spatial side net dimensions and (b) is the comparison of different temporal side net dimensions.} 
\label{dimension}
\end{figure}

\noindent 
$\bullet$ \textbf{Influence of different dimensions of spatial side net.}  
As the result shown in Fig.~\ref{dimension} (a), we conduct experiments on our proposed VTFPAR++ to analyze the impact of different dimensions of spatial side net. We can find that with the increasing dimension, the performance of our model continues to grow. However, this growth is not linear. The performance of the model peaks when the dimension is 336, but to ensure a lightweight architecture for the entire spatial side network, we finally choose 240 as the dimension for the spatial side network.

\noindent 
$\bullet$ \textbf{Influence of different dimensions of temporal side net.}
We can easily see from the result shown in Fig.~\ref{dimension} (b) that we determine the finalized temporal side network dimension of our VTFPAR++ model by experimenting with different dimensions. The interesting thing is that as our results were performed on the spatial side of the network, our experimental accuracy peaks when the dimension of the temporal side of the network is 336. Similarly, to ensure that our spatial side network is a lightweight architecture, we chose the 240 dimensions, which is slightly less expressive but more lightweight and easily fine-tunable.

\noindent 
$\bullet$ \textbf{Influence of different spatial interaction layers.} 
We show the results of the experiments we have done on different interaction layers on the spatial side of the network in Table~\ref{interaction}. In terms of the choice of spatial interaction levels, it can be found that more interaction levels lead to better performance, but this imposes a larger computational burden. We find that the performance of deep feature interactions is slightly improved compared to shallow CLIP~\cite{radford2021CLIP} features, and deep features are more discriminative than shallow ones. However, different attributes benefit from different features, so we choose to evenly select some layers from the CLIP features for interaction.

\begin{table}
\center
\small  
\caption{Comparing different interaction layers of spatial/temporal side network on MARs dataset } \label{interaction} 
\setlength{\tabcolsep}{1mm}{
\begin{tabular}{c|l|l|c}
\hline \toprule [0.5 pt]
\multicolumn{1}{c|}{Type} & \multicolumn{1}{c|}{Interaction type} & \multicolumn{1}{c|}{Interaction layers} &  \multicolumn{1}{c}{F1 score} \\  \hline
 \multirow{8}{*}{Spatial} & Ours & 0,3,6,9,11 & 83.22  \\ %基础实验
& Full-Layers  & 0,1,2,4,6,8,10,11 & \textbf{83.34}  \\ %7
& Shallowest-Layers  & 0,1,2,3,4 & 82.35  \\ %0
& Deepest-Layers  & 7,8,9,10,11 & 83.07\\ %1
& 4-Layers  & 3,6,9,11 &83.10  \\ %3
& 3-Layers  & 6,9,11 &83.03  \\ %4
& 2-Layers  & 9,11 &82.85  \\ %5
& 1-Layer & 11 &82.24  \\ %6
\hline
 \multirow{8}{*}{Temporal} & Ours & 0,3,6,9,11 & 83.22 \\ %基础实验
& Full-Layers & 0,1,2,4,6,8,10,11 & 83.02  \\ %7
& Shallowest-Layer & 0,1,2,3,4 & 82.79   \\ %0 
& Deepest-Layers & 7,8,9,10,11 & \textbf{83.31 }\\ %1
& 4-Layers & 3,6,9,11 &83.15  \\ %3
& 3-Layers & 6,9,11 &83.15  \\ %4
& 2-Layers & 9,11 &82.93  \\ %5
& 1-Layer & 11 &82.40  \\ %6
\hline \toprule [0.5 pt] 
\end{tabular} }
\end{table}

\noindent 
$\bullet$ \textbf{Influence of different temporal interaction layers.} As shown in Table~\ref{interaction}, the case of temporal interaction is slightly different from that of spatial interaction, and the performance is the best when we use the deepest 5 layers of CLIP visual features, with F1-score reaching 83.31, which exceeds the use of the shallowest layer of features by +0.52. Thus, the temporal interaction benefits more from the deeper features. Temporal features are essential for establishing inter-frame relationships and extracting robust pedestrian features. Relying on shallow CLIP visual features alone is insufficient for temporal modeling as they lack sufficient semantics.

\begin{table}
\center
\caption{Contrasting different spatial/temporal features aggregation methods on MARs dataset} 
\label{aggregation} 
\begin{tabular}{c|c|c}
\hline \toprule [0.5 pt]
\multicolumn{1}{c|}{\textbf{Module}} & \multicolumn{1}{c|}{\textbf{Fusion method}} &  \multicolumn{1}{c}{\textbf{F1 score}} \\  
\hline
\multirow{3}{*}{GRU}  & Spatio-Temporal & 82.36 \\
                      & Spatio     & 82.88 \\
                      & Temporal     & \textbf{83.17} \\ 
\hline
\multirow{3}{*}{LSTM} & Spatio-Temporal & 82.32 \\
                      & Spatio     & \textbf{83.04 }\\
                      & Temporal     & 82.54 \\ 
\hline
\multirow{3}{*}{MLP} & Spatio-Temporal &83.14  \\
                      & Spatio     &83.20  \\
                      & Temporal     &\textbf{83.27}  \\ 
\hline 
\multirow{1}{*}{GAP} & Spatio-Temporal &\textbf{83.22}   \\
\hline \toprule [0.5 pt] 
\end{tabular}
\end{table}

\noindent 
$\bullet$ \textbf{Analysis on various feature aggregation modules.} 
In order to determine the best module to fuse spatial and temporal features, we set different fusion modules in our proposed VTFPAR++ model. And we can find the result shown in Table~\ref{aggregation}, we set GRU~\cite{chung2014gru}, LSTM~\cite{hochreiter1997lstm}, MLP, and GAP as spatial and temporal feature fusion modules respectively. 
When GRU and MLP are selected as the feature aggregation module, it can be found that the effect of using temporal alone is better, and we analyze that this may because the temporal side net aggregates multi-level CLIP features across frames, and multi-level temporal features require more appropriate aggregation methods. Whereas the spatial features consist of multi-frame features, the inter-frame relationships have already been modeled by the temporal side net, thus, they only need to go through global average pooling. When both the temporal feature aggregation and spatial feature aggregation modules use learnable models, the lack of effective monitoring of the modeling labels makes it more difficult for the models to learn effective feature aggregation capabilities, leading to poor results.

\begin{table}
\center
\small  
\caption{ Comparison with other PEFT strategies on MARS dataset.}  
\label{otherPEFT} 
\begin{tabular}{c|ccc}
\hline \toprule [0.5 pt]
\multicolumn{1}{c|}{\multirow{1}{*}{Method}} & 
\multicolumn{1}{c}{Precision} & 
\multicolumn{1}{c}{Recall} & 
\multicolumn{1}{c}{F1 score} \\
\hline
LoRA                & 81.61 & 84.30 & 82.59 \\
Prompt-Tuning       & 82.03 & 82.66 & 81.74 \\
Adapter-Tuning      & 82.46 & 83.65 & 82.50 \\
Ours                & 82.27 & 84.87 & 83.22 \\
\hline \toprule [0.5 pt] 
\end{tabular} 
\end{table}

\begin{figure*}
\centering
\small
\includegraphics[width=7in]{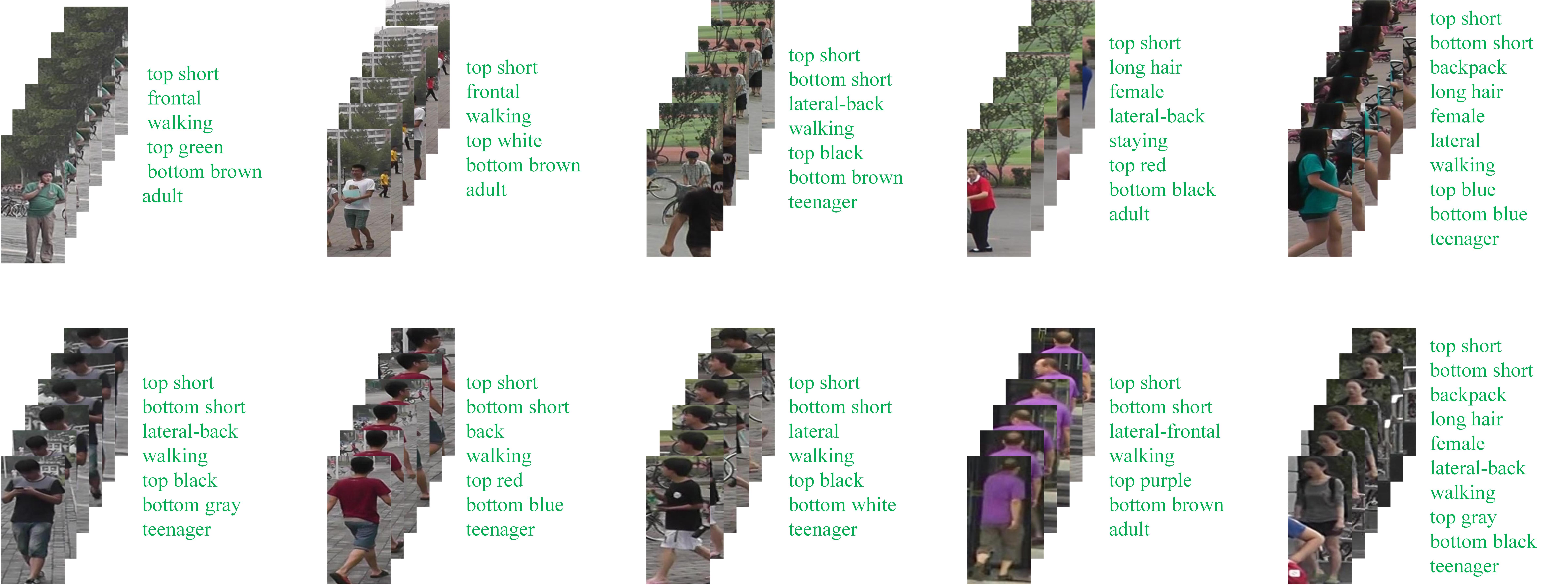}
\caption{Visualization of pedestrian attributes predicted by our proposed VTFPAR++ based on spatiotemporal side tuning. The \textcolor{SeaGreen4}{\emph{green}} attributes are corrected predicted ones.} 
\label{attResultsVIS} 
\end{figure*}

\begin{figure*}
\centering
\small
\includegraphics[width=7in]{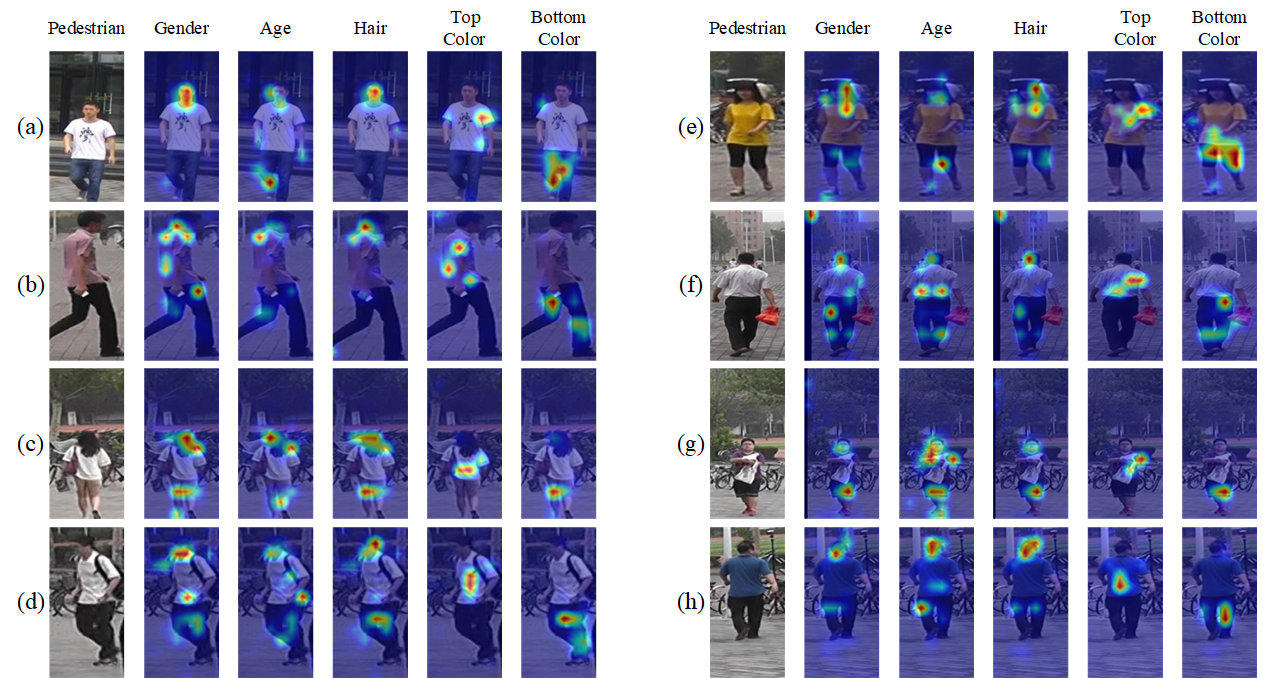}
\caption{Visualization of heat maps given the corresponding pedestrian attribute.} 
\label{heatmap} 
\end{figure*}

\begin{figure*}
\centering
\small
\includegraphics[width=7in]{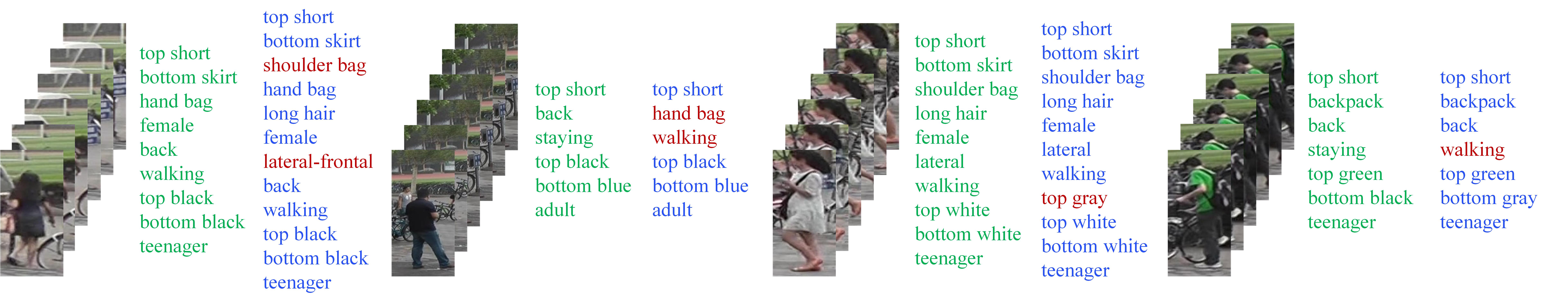}
\caption{Visualization of incorrect predicted attributes. \textcolor{SeaGreen4}{\emph{Green}} attributes are ground truth, the \textcolor{SlateBlue}{\emph{Blue}} (predicted correct attribute) and \textcolor{DarkRed}{\emph{Red}} (incorrect predicted attributes) attributes are predicted by our VTFPAR++. 
}  
\label{failedattResultsVIS} 
\end{figure*}

\noindent 
$\bullet$ \textbf{Comparison with other PEFT Strategies.} 
As shown in Table.~\ref{otherPEFT}, we compare the performance of different PEFT methods (i.e., Prompt tuning, Adapter, and LoRA)
based on the VTF~\cite{zhu2023videoCLIPPAR}, and find that our proposed VTFPAR++ significantly outperforms these strategies. 
Firstly, we compare with the Visual Prompt Tuning~\cite{2022vpt} by following PromptPAR~\cite{wang2023PromptPAR}~\footnote{\url{https://github.com/Event-AHU/OpenPAR}}, where to be fair we only included the prompt token in each layer of input to the visual branch of the CLIP~\cite{radford2021CLIP}, our method improves the precision, recall, and F1 score by +2.38, -0.31, and +1.16, respectively.
Our model also has significant performance gains over Adapter-Tuning~\cite{rebuffi2017learning}, i.e., the precision, recall, and F1 score are improved by -0.19, +1.22, and +0.72, respectively. 
We follow the example of LoRA-CLIP~\footnote{\url{https://github.com/jaisidhsingh/LoRA-CLIP}} by connecting a LoRA~\cite{hu2021lora} structure in parallel next to all linear layers of the CLIP visual encoder, and this approach results in three metrics that are lower by 0.66, 0.57, 0.63. 
Based on all these experimental results and comparisons, we can draw the conclusion that our proposed spatiotemporal works better than current state-of-the-art PEFT strategies.

\subsection{Visualization} \label{visualization}
In this section, we first give a case study demonstration of attributes that were predicted successfully on the MARS dataset. Immediately after, we also give some visualization results of attributes that were predicted unsuccessfully, and in order to show the prediction process of our model more intuitively, we also show a heatmap visualization of the predicted regions of interest of our VTFPAR++ model.

\noindent 
$\bullet$ \textbf{Attributes Predicted using Our VTFPAR++.}  
As shown in Fig.~\ref{attResultsVIS}, we provide 10 predictions of our model on the MARS dataset. It is easy to find that our proposed VTFPAR++ model has successfully predicted pedestrian attributes, such as top length, bottom length, motion, shoulder bag, etc.

\noindent 
$\bullet$ \textbf{Heatmap Visualization}
In order to show more intuitively the regions of interest in the process of predicting pedestrian attributes by our proposed model, we visualize a heat map of the model prediction process~\footnote{\url{https://github.com/jacobgil/vit-explain}}. As shown in Fig.~\ref{heatmap}, our model focuses on the correct region where the pedestrian attributes are located during the prediction process, e.g., the gender and age attributes focus on the pedestrian as a whole, the hair focus on the pedestrian top region, and the classification vectors for the upper and lower body color also correctly locate the pedestrian upper and lower body clothing. However, on some images, due to a series of problems such as occlusion, background noise, etc., the model's area of interest will also be concentrated on irrelevant parts, and we will work on improving this problem in our subsequent research.

\subsection{Failed Cases and Limitation Analysis} \label{failed}  %杰作
As illustrated in Fig~\ref{failedattResultsVIS}, we also give a visualization of the incorrectly predicted samples on the MARS dataset. From these samples, we can find that the recognition of the samples with multiple humans or videos containing background noise is very challenging. In our future works, we will consider designing novel target human localization networks to make our model understand which person we need to analyze. Also, we can further improve the backbone network to make it more friendly for practical employment. For example, the recently released state space model~\cite{Wang2024SSMSurvey} architecture performs better on the GPU memory consumption, which may be a promising research direction for video-based pedestrian attribute recognition. As the raw video contain about 60 frames, however, our current model only supports about 6 frames, which severely loses the raw spatial-temporal information.

\section{Conclusion and Future Works} \label{conclusion} 
In this paper, we formulate the video-based pedestrian attribute recognition as a video-text fusion problem and build a new framework, termed VTFPAR++, based on a pre-trained multi-modal foundation model CLIP. It mainly contains four sub-networks, including CLIP encoder, spatiotemporal side networks, multi-modal fusion Transformer, and attribute recognition head. We split, expand, and prompt the attribute into natural language descriptions and extract the semantic representations using a pre-trained CLIP text encoder. For the video input, we fix the pre-trained CLIP vision model and introduce the lightweight Transformer network as the side networks to achieve more efficient tuning. Then, we concatenate the dual modalities and feed them into a Transformer layer for vision-text fusion. Finally, we adopt the attribute recognition head for final prediction. Our experiments on two large-scale benchmark datasets fully validated the effectiveness of VTFPAR++. It also beats existing widely used parameter efficient fine tuning algorithms like the LoRA, Adapter, and Prompt tuning on the GPU memory usage, time cost, and recognition accuracy.

Although our framework performs well in these aspects, however, it also performs poor in some challenging scenarios. In addition, the complexity of the used Transformer layer is $\mathcal{O}(N^2)$, which is rather high for practical employment. In our future works, we will consider introducing more lightweight neural networks (e.g., State Space Model/Mamba~\cite{Wang2024SSMSurvey}) to replace the Transformer networks to further improve the efficiency of our model.

\small{ 
\bibliographystyle{IEEEtran}
\bibliography{reference}

\begin{thebibliography}{10}
\providecommand{\url}[1]{#1}
\csname url@samestyle\endcsname
\providecommand{\newblock}{\relax}
\providecommand{\bibinfo}[2]{#2}
\providecommand{\BIBentrySTDinterwordspacing}{\spaceskip=0pt\relax}
\providecommand{\BIBentryALTinterwordstretchfactor}{4}
\providecommand{\BIBentryALTinterwordspacing}{\spaceskip=\fontdimen2\font plus
\BIBentryALTinterwordstretchfactor\fontdimen3\font minus
  \fontdimen4\font\relax}
\providecommand{\BIBforeignlanguage}[2]{{%
\expandafter\ifx\csname l@#1\endcsname\relax
\typeout{** WARNING: IEEEtran.bst: No hyphenation pattern has been}%
\typeout{** loaded for the language `#1'. Using the pattern for}%
\typeout{** the default language instead.}%
\else
\language=\csname l@#1\endcsname
\fi
#2}}
\providecommand{\BIBdecl}{\relax}
\BIBdecl

\bibitem{wang2022PARsurvey}
X.~Wang, S.~Zheng, R.~Yang, A.~Zheng, Z.~Chen, J.~Tang, and B.~Luo,
  ``Pedestrian attribute recognition: A survey,'' \emph{Pattern Recognition},
  vol. 121, p. 108220, 2022.

\bibitem{cheng2022VTB}
X.~Cheng, M.~Jia, Q.~Wang, and J.~Zhang, ``A simple visual-textual baseline for
  pedestrian attribute recognition,'' \emph{IEEE Transactions on Circuits and
  Systems for Video Technology}, vol.~32, no.~10, pp. 6994--7004, 2022.

\bibitem{he2016resnet}
K.~He, X.~Zhang, S.~Ren, and J.~Sun, ``Deep residual learning for image
  recognition,'' in \emph{Proceedings of the IEEE conference on computer vision
  and pattern recognition}, 2016, pp. 770--778.

\bibitem{chung2014gru}
J.~Chung, C.~Gulcehre, K.~Cho, and Y.~Bengio, ``Empirical evaluation of gated
  recurrent neural networks on sequence modeling,'' \emph{arXiv preprint
  arXiv:1412.3555}, 2014.

\bibitem{wang2017RNNPAR}
J.~Wang, X.~Zhu, S.~Gong, and W.~Li, ``Attribute recognition by joint recurrent
  learning of context and correlation,'' in \emph{Proceedings of the IEEE
  International Conference on Computer Vision}, 2017, pp. 531--540.

\bibitem{hochreiter1997lstm}
S.~Hochreiter and J.~Schmidhuber, ``Long short-term memory,'' \emph{Neural
  computation}, vol.~9, no.~8, pp. 1735--1780, 1997.

\bibitem{vaswani2017Transformer}
A.~Vaswani, N.~Shazeer, N.~Parmar, J.~Uszkoreit, L.~Jones, A.~N. Gomez,
  L.~Kaiser, and I.~Polosukhin, ``Attention is all you need,'' in
  \emph{Proceedings of the 31st International Conference on Neural Information
  Processing Systems}, 2017, pp. 6000–--6010.

\bibitem{DosovitskiyViT}
A.~Dosovitskiy, L.~Beyer, A.~Kolesnikov, D.~Weissenborn, X.~Zhai,
  T.~Unterthiner, M.~Dehghani, M.~Minderer, G.~Heigold, S.~Gelly, J.~Uszkoreit,
  and N.~Houlsby, ``An image is worth 16x16 words: Transformers for image
  recognition at scale,'' in \emph{International Conference on Learning
  Representations}, 2021.

\bibitem{wang2023MMPTMs}
X.~Wang, G.~Chen, G.~Qian, P.~Gao, X.-Y. Wei, Y.~Wang, Y.~Tian, and W.~Gao,
  ``Large-scale multi-modal pre-trained models: A comprehensive survey,''
  \emph{Machine Intelligence Research}, vol.~20, no.~4, pp. 447--482, 2023.

\bibitem{zhao2023transformerVLT}
H.~Zhao, X.~Wang, D.~Wang, H.~Lu, and X.~Ruan, ``Transformer vision-language
  tracking via proxy token guided cross-modal fusion,'' \emph{Pattern
  Recognition Letters}, vol. 168, pp. 10--16, 2023.

\bibitem{wang2021TNL2K}
X.~Wang, X.~Shu, Z.~Zhang, B.~Jiang, Y.~Wang, Y.~Tian, and F.~Wu, ``Towards
  more flexible and accurate object tracking with natural language: Algorithms
  and benchmark,'' in \emph{Proceedings of the IEEE/CVF Conference on Computer
  Vision and Pattern Recognition}, 2021, pp. 13\,763--13\,773.

\bibitem{tang2022drformer}
Z.~Tang and J.~Huang, ``Drformer: Learning dual relations using transformer for
  pedestrian attribute recognition,'' \emph{Neurocomputing}, vol. 497, pp.
  159--169, 2022.

\bibitem{zhang2020attributedetection}
J.~Zhang, L.~Lin, J.~Zhu, Y.~Li, Y.-c. Chen, Y.~Hu, and S.~C. Hoi,
  ``Attribute-aware pedestrian detection in a crowd,'' \emph{IEEE Transactions
  on Multimedia}, vol.~23, pp. 3085--3097, 2020.

\bibitem{li2023attmot}
Y.~Li, Z.~Xiao, L.~Yang, D.~Meng, X.~Zhou, H.~Fan, and L.~Zhang, ``Attmot:
  Improving multiple-object tracking by introducing auxiliary pedestrian
  attributes,'' \emph{arXiv preprint arXiv:2308.07537}, 2023.

\bibitem{zheng2022progCMreid}
A.~Zheng, P.~Pan, H.~Li, C.~Li, B.~Luo, C.~Tan, and R.~Jia, ``Progressive
  attribute embedding for accurate cross-modality person re-id,'' in
  \emph{Proceedings of the 30th ACM International Conference on Multimedia},
  2022, pp. 4309--4317.

\bibitem{chen2019videoPAR}
Z.~Chen, A.~Li, and Y.~Wang, ``A temporal attentive approach for video-based
  pedestrian attribute recognition,'' in \emph{Pattern Recognition and Computer
  Vision: Second Chinese Conference, PRCV 2019, Xi’an, China, November 8--11,
  2019, Proceedings, Part II 2}.\hskip 1em plus 0.5em minus 0.4em\relax
  Springer, 2019, pp. 209--220.

\bibitem{radford2021CLIP}
A.~Radford, J.~W. Kim, C.~Hallacy, A.~Ramesh, G.~Goh, S.~Agarwal, G.~Sastry,
  A.~Askell, P.~Mishkin, J.~Clark \emph{et~al.}, ``Learning transferable visual
  models from natural language supervision,'' in \emph{International Conference
  on Machine Learning}.\hskip 1em plus 0.5em minus 0.4em\relax PMLR, 2021, pp.
  8748--8763.

\bibitem{wu2018exploit}
Y.~Wu, Y.~Lin, X.~Dong, Y.~Yan, W.~Ouyang, and Y.~Yang, ``Exploit the unknown
  gradually: One-shot video-based person re-identification by stepwise
  learning,'' in \emph{Proceedings of the IEEE conference on computer vision
  and pattern recognition}, 2018, pp. 5177--5186.

\bibitem{abdulnabi2015multi}
A.~H. Abdulnabi, G.~Wang, J.~Lu, and K.~Jia, ``Multi-task cnn model for
  attribute prediction,'' \emph{IEEE Transactions on Multimedia}, vol.~17,
  no.~11, pp. 1949--1959, 2015.

\bibitem{zhang2014panda}
N.~Zhang, M.~Paluri, M.~Ranzato, T.~Darrell, and L.~Bourdev, ``Panda: Pose
  aligned networks for deep attribute modeling,'' in \emph{Proceedings of the
  IEEE conference on computer vision and pattern recognition}, 2014, pp.
  1637--1644.

\bibitem{wang2016cnn}
J.~Wang, Y.~Yang, J.~Mao, Z.~Huang, C.~Huang, and W.~Xu, ``Cnn-rnn: A unified
  framework for multi-label image classification,'' in \emph{Proceedings of the
  IEEE conference on computer vision and pattern recognition}, 2016, pp.
  2285--2294.

\bibitem{tian2015pedestrian}
Y.~Tian, P.~Luo, X.~Wang, and X.~Tang, ``Pedestrian detection aided by deep
  learning semantic tasks,'' in \emph{Proceedings of the IEEE conference on
  computer vision and pattern recognition}, 2015, pp. 5079--5087.

\bibitem{li2019visual}
Q.~Li, X.~Zhao, R.~He, and K.~Huang, ``Visual-semantic graph reasoning for
  pedestrian attribute recognition,'' in \emph{Proceedings of the AAAI
  conference on artificial intelligence}, vol.~33, no.~01, 2019, pp.
  8634--8641.

\bibitem{park2017attribute}
S.~Park, B.~X. Nie, and S.-C. Zhu, ``Attribute and-or grammar for joint parsing
  of human pose, parts and attributes,'' \emph{IEEE transactions on pattern
  analysis and machine intelligence}, vol.~40, no.~7, pp. 1555--1569, 2017.

\bibitem{specker2020evaluation}
A.~Specker, A.~Schumann, and J.~Beyerer, ``An evaluation of design choices for
  pedestrian attribute recognition in video,'' in \emph{2020 IEEE International
  Conference on Image Processing (ICIP)}.\hskip 1em plus 0.5em minus
  0.4em\relax IEEE, 2020, pp. 2331--2335.

\bibitem{fan2023parformer}
X.~Fan, Y.~Zhang, Y.~Lu, and H.~Wang, ``Parformer: transformer-based multi-task
  network for pedestrian attribute recognition,'' \emph{IEEE Transactions on
  Circuits and Systems for Video Technology}, 2023.

\bibitem{lee2021robust}
G.~Lee, K.~Yun, and J.~Cho, ``Robust pedestrian attribute recognition using
  group sparsity for occlusion videos,'' \emph{arXiv preprint
  arXiv:2110.08708}, 2021.

\bibitem{thakare2024let}
K.~V. Thakare, D.~P. Dogra, H.~Choi, H.~Kim, and I.-J. Kim, ``Let's observe
  them over time: An improved pedestrian attribute recognition approach,'' in
  \emph{Proceedings of the IEEE/CVF Winter Conference on Applications of
  Computer Vision}, 2024, pp. 708--717.

\bibitem{liu2024pedestrian}
Z.~Liu, D.~Li, X.~Zhang, Z.~Zhang, P.~Zhang, C.~Shan, and J.~Han, ``Pedestrian
  attribute recognition via spatio-temporal relationship learning for visual
  surveillance,'' \emph{ACM Transactions on Multimedia Computing,
  Communications and Applications}, vol.~20, no.~6, pp. 1--15, 2024.

\bibitem{devlin2018bert}
J.~Devlin, M.-W. Chang, K.~Lee, and K.~Toutanova, ``Bert: Pre-training of deep
  bidirectional transformers for language understanding,'' \emph{arXiv preprint
  arXiv:1810.04805}, 2018.

\bibitem{radford2019language}
A.~Radford, J.~Wu, R.~Child, D.~Luan, D.~Amodei, I.~Sutskever \emph{et~al.},
  ``Language models are unsupervised multitask learners,'' \emph{OpenAI blog},
  vol.~1, no.~8, p.~9, 2019.

\bibitem{brown2020language}
T.~Brown, B.~Mann, N.~Ryder, M.~Subbiah, J.~D. Kaplan, P.~Dhariwal,
  A.~Neelakantan, P.~Shyam, G.~Sastry, A.~Askell \emph{et~al.}, ``Language
  models are few-shot learners,'' \emph{Advances in neural information
  processing systems}, vol.~33, pp. 1877--1901, 2020.

\bibitem{touvron2023llama}
H.~Touvron, T.~Lavril, G.~Izacard, X.~Martinet, M.-A. Lachaux, T.~Lacroix,
  B.~Rozi{\`e}re, N.~Goyal, E.~Hambro, F.~Azhar \emph{et~al.}, ``Llama: Open
  and efficient foundation language models,'' \emph{arXiv preprint
  arXiv:2302.13971}, 2023.

\bibitem{thoppilan2022lamda}
R.~Thoppilan, D.~De~Freitas, J.~Hall, N.~Shazeer, A.~Kulshreshtha, H.-T. Cheng,
  A.~Jin, T.~Bos, L.~Baker, Y.~Du \emph{et~al.}, ``Lamda: Language models for
  dialog applications,'' \emph{arXiv preprint arXiv:2201.08239}, 2022.

\bibitem{yang2023baichuan}
A.~Yang, B.~Xiao, B.~Wang, B.~Zhang, C.~Bian, C.~Yin, C.~Lv, D.~Pan, D.~Wang,
  D.~Yan \emph{et~al.}, ``Baichuan 2: Open large-scale language models,''
  \emph{arXiv preprint arXiv:2309.10305}, 2023.

\bibitem{jia2021scaling}
C.~Jia, Y.~Yang, Y.~Xia, Y.-T. Chen, Z.~Parekh, H.~Pham, Q.~Le, Y.-H. Sung,
  Z.~Li, and T.~Duerig, ``Scaling up visual and vision-language representation
  learning with noisy text supervision,'' in \emph{International conference on
  machine learning}.\hskip 1em plus 0.5em minus 0.4em\relax PMLR, 2021, pp.
  4904--4916.

\bibitem{tan2019lxmert}
H.~Tan and M.~Bansal, ``Lxmert: Learning cross-modality encoder representations
  from transformers,'' \emph{arXiv preprint arXiv:1908.07490}, 2019.

\bibitem{lu2019vilbert}
J.~Lu, D.~Batra, D.~Parikh, and S.~Lee, ``Vilbert: Pretraining task-agnostic
  visiolinguistic representations for vision-and-language tasks,''
  \emph{Advances in neural information processing systems}, vol.~32, 2019.

\bibitem{kirillov2023SAM}
A.~Kirillov, E.~Mintun, N.~Ravi, H.~Mao, C.~Rolland, L.~Gustafson, T.~Xiao,
  S.~Whitehead, A.~C. Berg, W.-Y. Lo \emph{et~al.}, ``Segment anything,'' in
  \emph{Proceedings of the IEEE/CVF International Conference on Computer
  Vision}, 2023, pp. 4015--4026.

\bibitem{radford2021learning}
A.~Radford, J.~W. Kim, C.~Hallacy, A.~Ramesh, G.~Goh, S.~Agarwal, G.~Sastry,
  A.~Askell, P.~Mishkin, J.~Clark \emph{et~al.}, ``Learning transferable visual
  models from natural language supervision,'' in \emph{International conference
  on machine learning}.\hskip 1em plus 0.5em minus 0.4em\relax PMLR, 2021, pp.
  8748--8763.

\bibitem{wang2023PromptPAR}
X.~Wang, J.~Jin, C.~Li, J.~Tang, C.~Zhang, and W.~Wang, ``Pedestrian attribute
  recognition via clip based prompt vision-language fusion,'' \emph{arXiv
  preprint arXiv:2312.10692}, 2023.

\bibitem{jin2023sequencepar}
J.~Jin, X.~Wang, C.~Li, L.~Huang, and J.~Tang, ``Sequencepar: Understanding
  pedestrian attributes via a sequence generation paradigm,'' \emph{arXiv
  preprint arXiv:2312.01640}, 2023.

\bibitem{wang2023hulk}
Y.~Wang, Y.~Wu, S.~Tang, W.~He, X.~Guo, F.~Zhu, L.~Bai, R.~Zhao, J.~Wu, T.~He
  \emph{et~al.}, ``Hulk: A universal knowledge translator for human-centric
  tasks,'' \emph{arXiv preprint arXiv:2312.01697}, 2023.

\bibitem{yuan2024hap}
J.~Yuan, X.~Zhang, H.~Zhou, J.~Wang, Z.~Qiu, Z.~Shao, S.~Zhang, S.~Long,
  K.~Kuang, K.~Yao \emph{et~al.}, ``Hap: Structure-aware masked image modeling
  for human-centric perception,'' \emph{Advances in Neural Information
  Processing Systems}, vol.~36, 2024.

\bibitem{zuo2023plip}
J.~Zuo, C.~Yu, N.~Sang, and C.~Gao, ``Plip: Language-image pre-training for
  person representation learning,'' \emph{arXiv preprint arXiv:2305.08386},
  2023.

\bibitem{zhu2023videoCLIPPAR}
J.~Zhu, J.~Jin, Z.~Yang, X.~Wu, and X.~Wang, ``Learning clip guided visual-text
  fusion transformer for video-based pedestrian attribute recognition,'' in
  \emph{2023 IEEE/CVF Conference on Computer Vision and Pattern Recognition
  Workshops (CVPRW)}.\hskip 1em plus 0.5em minus 0.4em\relax IEEE, 2023, pp.
  2626--2629.

\bibitem{dong2022survey}
Q.~Dong, L.~Li, D.~Dai, C.~Zheng, Z.~Wu, B.~Chang, X.~Sun, J.~Xu, and Z.~Sui,
  ``A survey on in-context learning,'' \emph{arXiv preprint arXiv:2301.00234},
  2022.

\bibitem{zhang2023instruction}
S.~Zhang, L.~Dong, X.~Li, S.~Zhang, X.~Sun, S.~Wang, J.~Li, R.~Hu, T.~Zhang,
  F.~Wu \emph{et~al.}, ``Instruction tuning for large language models: A
  survey,'' \emph{arXiv preprint arXiv:2308.10792}, 2023.

\bibitem{wei2022chain}
J.~Wei, X.~Wang, D.~Schuurmans, M.~Bosma, F.~Xia, E.~Chi, Q.~V. Le, D.~Zhou
  \emph{et~al.}, ``Chain-of-thought prompting elicits reasoning in large
  language models,'' \emph{Advances in neural information processing systems},
  vol.~35, pp. 24\,824--24\,837, 2022.

\bibitem{2022vpt}
M.~Jia, L.~Tang, B.-C. Chen, C.~Cardie, S.~Belongie, B.~Hariharan, and S.-N.
  Lim, ``Visual prompt tuning,'' in \emph{European Conference on Computer
  Vision}.\hskip 1em plus 0.5em minus 0.4em\relax Springer, 2022, pp. 709--727.

\bibitem{zhou2022coop}
K.~Zhou, J.~Yang, C.~C. Loy, and Z.~Liu, ``Learning to prompt for
  vision-language models,'' \emph{International Journal of Computer Vision},
  vol. 130, no.~9, pp. 2337--2348, 2022.

\bibitem{zhou2022cocoop}
{K. Zhou, J. Yang, CC. Loy, and Z. Liu}, ``Conditional prompt learning for
  vision-language models,'' in \emph{Proceedings of the IEEE/CVF Conference on
  Computer Vision and Pattern Recognition}, 2022, pp. 16\,816--16\,825.

\bibitem{rebuffi2017learning}
S.-A. Rebuffi, H.~Bilen, and A.~Vedaldi, ``Learning multiple visual domains
  with residual adapters,'' in \emph{Proceedings of the 31st International
  Conference on Neural Information Processing Systems}, 2017, pp. 506--516.

\bibitem{houlsby2019parameter}
N.~Houlsby, A.~Giurgiu, S.~Jastrzebski, B.~Morrone, Q.~De~Laroussilhe,
  A.~Gesmundo, M.~Attariyan, and S.~Gelly, ``Parameter-efficient transfer
  learning for nlp,'' in \emph{International conference on machine
  learning}.\hskip 1em plus 0.5em minus 0.4em\relax PMLR, 2019, pp. 2790--2799.

\bibitem{gao2024clipadapterc}
P.~Gao, S.~Geng, R.~Zhang, T.~Ma, R.~Fang, Y.~Zhang, H.~Li, and Y.~Qiao,
  ``Clip-adapter: Better vision-language models with feature adapters,''
  \emph{International Journal of Computer Vision}, vol. 132, no.~2, pp.
  581--595, 2024.

\bibitem{hu2021lora}
E.~J. Hu, Y.~Shen, P.~Wallis, Z.~Allen-Zhu, Y.~Li, S.~Wang, L.~Wang, and
  W.~Chen, ``Lora: Low-rank adaptation of large language models,'' \emph{arXiv
  preprint arXiv:2106.09685}, 2021.

\bibitem{dou2023loramoe}
S.~Dou, E.~Zhou, Y.~Liu, S.~Gao, J.~Zhao, W.~Shen, Y.~Zhou, Z.~Xi, X.~Wang,
  X.~Fan \emph{et~al.}, ``Loramoe: Revolutionizing mixture of experts for
  maintaining world knowledge in language model alignment,'' \emph{arXiv
  preprint arXiv:2312.09979}, 2023.

\bibitem{jacobs1991adaptive}
R.~A. Jacobs, M.~I. Jordan, S.~J. Nowlan, and G.~E. Hinton, ``Adaptive mixtures
  of local experts,'' \emph{Neural computation}, vol.~3, no.~1, pp. 79--87,
  1991.

\bibitem{zhang2020side}
J.~O. Zhang, A.~Sax, A.~Zamir, L.~Guibas, and J.~Malik, ``Side-tuning: a
  baseline for network adaptation via additive side networks,'' in
  \emph{Computer Vision--ECCV 2020: 16th European Conference, Glasgow, UK,
  August 23--28, 2020, Proceedings, Part III 16}.\hskip 1em plus 0.5em minus
  0.4em\relax Springer, 2020, pp. 698--714.

\bibitem{sung2022lst}
Y.-L. Sung, J.~Cho, and M.~Bansal, ``Lst: Ladder side-tuning for parameter and
  memory efficient transfer learning,'' \emph{Advances in Neural Information
  Processing Systems}, vol.~35, pp. 12\,991--13\,005, 2022.

\bibitem{ji20123d}
S.~Ji, W.~Xu, M.~Yang, and K.~Yu, ``3d convolutional neural networks for human
  action recognition,'' \emph{IEEE transactions on pattern analysis and machine
  intelligence}, vol.~35, no.~1, pp. 221--231, 2012.

\bibitem{mclaughlin2016recurrent}
N.~McLaughlin, J.~M. Del~Rincon, and P.~Miller, ``Recurrent convolutional
  network for video-based person re-identification,'' in \emph{Proceedings of
  the IEEE conference on computer vision and pattern recognition}, 2016, pp.
  1325--1334.

\bibitem{tang2019improving}
C.~Tang, L.~Sheng, Z.~Zhang, and X.~Hu, ``Improving pedestrian attribute
  recognition with weakly-supervised multi-scale attribute-specific
  localization,'' in \emph{Proceedings of the IEEE/CVF International Conference
  on Computer Vision}, 2019, pp. 4997--5006.

\bibitem{jia2021spatial}
J.~Jia, X.~Chen, and K.~Huang, ``Spatial and semantic consistency
  regularizations for pedestrian attribute recognition,'' in \emph{Proceedings
  of the IEEE/CVF international conference on computer vision}, 2021, pp.
  962--971.

\bibitem{zhao2023tra}
Y.~Zhao, H.~Jin, X.~Shi, and H.~Sun, ``Temporal related attention for
  video-based pedestrian attribute recognition,'' in \emph{2023 18th
  International Conference on Intelligent Systems and Knowledge Engineering
  (ISKE)}.\hskip 1em plus 0.5em minus 0.4em\relax IEEE, 2023, pp. 93--97.

\bibitem{chen2019temporal}
Z.~Chen, A.~Li, and Y.~Wang, ``A temporal attentive approach for video-based
  pedestrian attribute recognition,'' in \emph{Pattern Recognition and Computer
  Vision: Second Chinese Conference, PRCV 2019, Xi’an, China, November 8--11,
  2019, Proceedings, Part II 2}.\hskip 1em plus 0.5em minus 0.4em\relax
  Springer, 2019, pp. 209--220.

\bibitem{zheng2016mars}
L.~Zheng, Z.~Bie, Y.~Sun, J.~Wang, C.~Su, S.~Wang, and Q.~Tian, ``Mars: A video
  benchmark for large-scale person re-identification,'' in \emph{Computer
  Vision--ECCV 2016: 14th European Conference, Amsterdam, The Netherlands,
  October 11-14, 2016, Proceedings, Part VI 14}.\hskip 1em plus 0.5em minus
  0.4em\relax Springer, 2016, pp. 868--884.

\bibitem{ristani2016performance}
E.~Ristani, F.~Solera, R.~Zou, R.~Cucchiara, and C.~Tomasi, ``Performance
  measures and a data set for multi-target, multi-camera tracking,'' in
  \emph{European conference on computer vision}.\hskip 1em plus 0.5em minus
  0.4em\relax Springer, 2016, pp. 17--35.

\bibitem{kingma2014adam}
D.~P. Kingma and J.~Ba, ``Adam: {A} method for stochastic optimization,'' in
  \emph{{ICLR} (Poster)}, 2015.

\bibitem{paszke2019pytorch}
A.~Paszke, S.~Gross, F.~Massa, A.~Lerer, J.~Bradbury, G.~Chanan, T.~Killeen,
  Z.~Lin, N.~Gimelshein, L.~Antiga \emph{et~al.}, ``Pytorch: An imperative
  style, high-performance deep learning library,'' \emph{Advances in neural
  information processing systems}, vol.~32, 2019.

\bibitem{Wang2024SSMSurvey}
X.~Wang, S.~Wang, Y.~Ding, Y.~Li, W.~Wu, Y.~Rong, W.~Kong, J.~Huang, S.~Li,
  H.~Yang, Z.~Wang, B.~Jiang, C.~Li, Y.~Wang, Y.~Tian, and J.~Tang, ``State
  space model for new-generation network alternative to transformers: A
  survey,'' 2024.

\end{thebibliography}
}

% a masterpiece is born
\end{document}